\documentclass[10pt,twocolumn,letterpaper]{article}

\usepackage{iccv}
\usepackage{times}
\usepackage{epsfig}
\usepackage{graphicx}
\usepackage{amsmath}
\usepackage{amssymb}

\usepackage{listings}
\usepackage{color}
\usepackage{subcaption}
\usepackage{mathtools}
\usepackage{paralist} 
\usepackage{colortbl}
\usepackage{nicefrac}
\usepackage{multirow}
\usepackage[table,dvipsnames]{xcolor}
\usepackage{makecell}
\usepackage[outline]{contour}
\usepackage{soul}

\usepackage[percent]{overpic}
\usepackage{xcolor}

\usepackage{pifont}
\definecolor{redc}{rgb}{1.,0,0}
\definecolor{greenc}{rgb}{0,.8,0}
\newcommand{\cmark}{\checkmark}%
\newcommand{\xmark}{}%

\usepackage{appendix}

\usepackage[pagebackref=true,breaklinks=true,letterpaper=true,colorlinks,bookmarks=false]{hyperref}

\iccvfinalcopy 

\def\iccvPaperID{8765} 

\begin{document}

\definecolor{yellow}{rgb}{1,1, 0.6}
\definecolor{lightyellow}{rgb}{1,1, 0.8}
\definecolor{orange}{rgb}{1, 0.8, 0.6}
\definecolor{red}{rgb}{1, 0.6, 0.6}

\definecolor{wincolor}{rgb}{0.85, 0.0, 0.0}

\definecolor{darkyellow}{rgb}{0.8, 0.8, 0.5}
\definecolor{darkred}{rgb}{0.7, 0.3, 0.3}
\definecolor{darkgreen}{rgb}{0.3, 0.7, 0.3}
\definecolor{blue}{rgb}{0, 0, 1.0}
\definecolor{green}{rgb}{0, 1.0, 0}
\definecolor{pink}{rgb}{1, 0.4, 0.7}

\newcommand{\barron}[1]{{\color{blue} barron: #1}}
\newcommand{\hedman}[1]{{\color{orange} hedman: #1}}
\newcommand{\todo}[1]{{\color{pink} TODO: #1}}

\newcommand{\mbf}[1]{{\mathbf{#1}}}

\let\originalleft\left
\let\originalright\right
\renewcommand{\left}{\mathopen{}\mathclose\bgroup\originalleft}
\renewcommand{\right}{\aftergroup\egroup\originalright}

\newcommand{\norm}[1]{\left\lVert#1\right\rVert}

\newcommand{\expo}[1]{\exp\left(#1\right)}

\newcommand{\modeltheta}{\mathrm{\Theta}}
\newcommand{\absrp}{\sigma}

\newcommand{\numsamples}{N}
\newcommand{\numsamplescoarse}{N_c}
\newcommand{\numsamplesfine}{N_f}
\newcommand{\timenear}{t_n}
\newcommand{\timefar}{t_f}
\newcommand{\deltatime}{\delta}

\newcommand{\posxy}{xy}
\newcommand{\posxyz}{xyz}
\newcommand{\angletheta}{\theta}
\newcommand{\anglephi}{\phi}
\newcommand{\posall}{\posxyz\angletheta\anglephi}

\newcommand{\numfrequencies}{L}

\newcommand{\Ltrain}{\mathcal{L}}
\newcommand{\raybatch}{\mathcal{R}}
\newcommand{\Ccoarse}{\hat{C}_c(\ray)}
\newcommand{\Cfine}{\hat{C}_f(\ray)}
\newcommand{\Ctrue}{C(\ray)}
\newcommand{\pweight}{w}
\newcommand{\normpweight}{\hat{w}}

\newcommand{\scenename}[1]{\textit{#1}}

\newcommand{\shortpara}[1]{\noindent {\bf #1}\,\,\,}


%
%

\newcommand{\posenc}{Positional Encoding\xspace}
\newcommand{\shortposenc}{PE\xspace}

\newcommand{\numhidden}{n}

\newcommand{\posencfun}{\gamma}
\newcommand{\coordenc}{\gamma(\mathbf{x})}

\newcommand{\rayorigin}{\mathbf{o}}
\newcommand{\raydir}{\mathbf{d}}
\newcommand{\transpose}{{\operatorname{T}}}

\newcommand\estimate[1]{\hat{#1}}
\newcommand{\ray}{\mathbf{r}}
\newcommand{\normsq}[1]{\big\lVert#1\big\rVert^2_2}
\newcommand{\position}{\mathbf{x}}
\newcommand{\col}{\mathbf{c}}
\newcommand{\feat}{\mathbf{v}_s}
\newcommand{\Col}{\mathbf{C}}
\newcommand{\Feat}{\mathbf{V}_s}
\newcommand{\truecol}{\mathbf{C}}
\newcommand{\mlp}{\operatorname{MLP}}
\newcommand{\modelweights}{\Theta}
\newcommand{\modelweightsspec}{\Phi}
\newcommand\decay[1]{\alpha \left( #1 \right)}

\newcommand{\textpyr}[5]{
	\begin{overpic}[width=1.3in]{figures/multiblender_pyrs/#1}
	\put (67,2) {\tiny \sethlcolor{white}\hl{$#2$}}
	\put (84,29) {\tiny \sethlcolor{white}\hl{$#3$}}
	\put (84,18) {\tiny \sethlcolor{white}\hl{$#4$}}
	\put (76,10) {\tiny \sethlcolor{white}\hl{$#5$}}
    \end{overpic}
}

\newcommand{\flatim}[1]{
    \includegraphics[width=0.9in]{figures/multiblender_pyrs/#1}
}

\title{Baking Neural Radiance Fields for Real-Time View Synthesis}


\author{
Peter Hedman 
\quad
Pratul P. Srinivasan 
\quad 
Ben Mildenhall
\quad
Jonathan T. Barron
\quad
Paul Debevec
 \\\\
{Google Research}
}

\maketitle

\begin{abstract}
Neural volumetric representations such as Neural Radiance Fields (NeRF) have emerged as a compelling technique for learning to represent 3D scenes from images with the goal of rendering photorealistic images of the scene from unobserved viewpoints. However, NeRF's computational requirements are prohibitive for real-time applications: rendering views from a trained NeRF requires querying a multilayer perceptron (MLP) hundreds of times \emph{per ray}.
We present a method to train a NeRF, then precompute and store (i.e. ``bake'') it as a novel representation called a Sparse Neural Radiance Grid (SNeRG) that enables real-time rendering on commodity hardware.
To achieve this, we introduce 1) a reformulation of NeRF's architecture, and 2) a sparse voxel grid representation with learned feature vectors. The resulting scene representation retains NeRF's ability to render fine geometric details and view-dependent appearance, is compact (averaging less than 90 MB per scene), and can be rendered in real-time (higher than 30 frames per second on a laptop GPU). Actual screen captures are shown in our video.

\end{abstract}

\section{Introduction}

The task of view synthesis --- using observed images to recover a 3D scene representation that can render the scene from novel unobserved viewpoints --- has recently seen dramatic progress as a result of using neural volumetric representations. In particular, Neural Radiance Fields (NeRF)~\cite{mildenhall20} are able to render photorealistic novel views with fine geometric details and realistic view-dependent appearance by representing a scene as a continuous volumetric function, parameterized by a multilayer perceptron (MLP) that maps from a continuous 3D position to the volume density and view-dependent emitted radiance at that location. Unfortunately, NeRF's rendering procedure is quite slow: rendering a ray requires querying an MLP hundreds of times, such that rendering a frame at $800\times800$ resolution takes roughly a minute on a modern GPU. This prevents NeRF from being used for interactive view synthesis applications such as virtual and augmented reality, or even simply inspecting a recovered 3D model in a web browser.

\begin{figure}[!t]
    \begin{minipage}{\linewidth}%
      \centering%
      \includegraphics[width=1.\columnwidth]{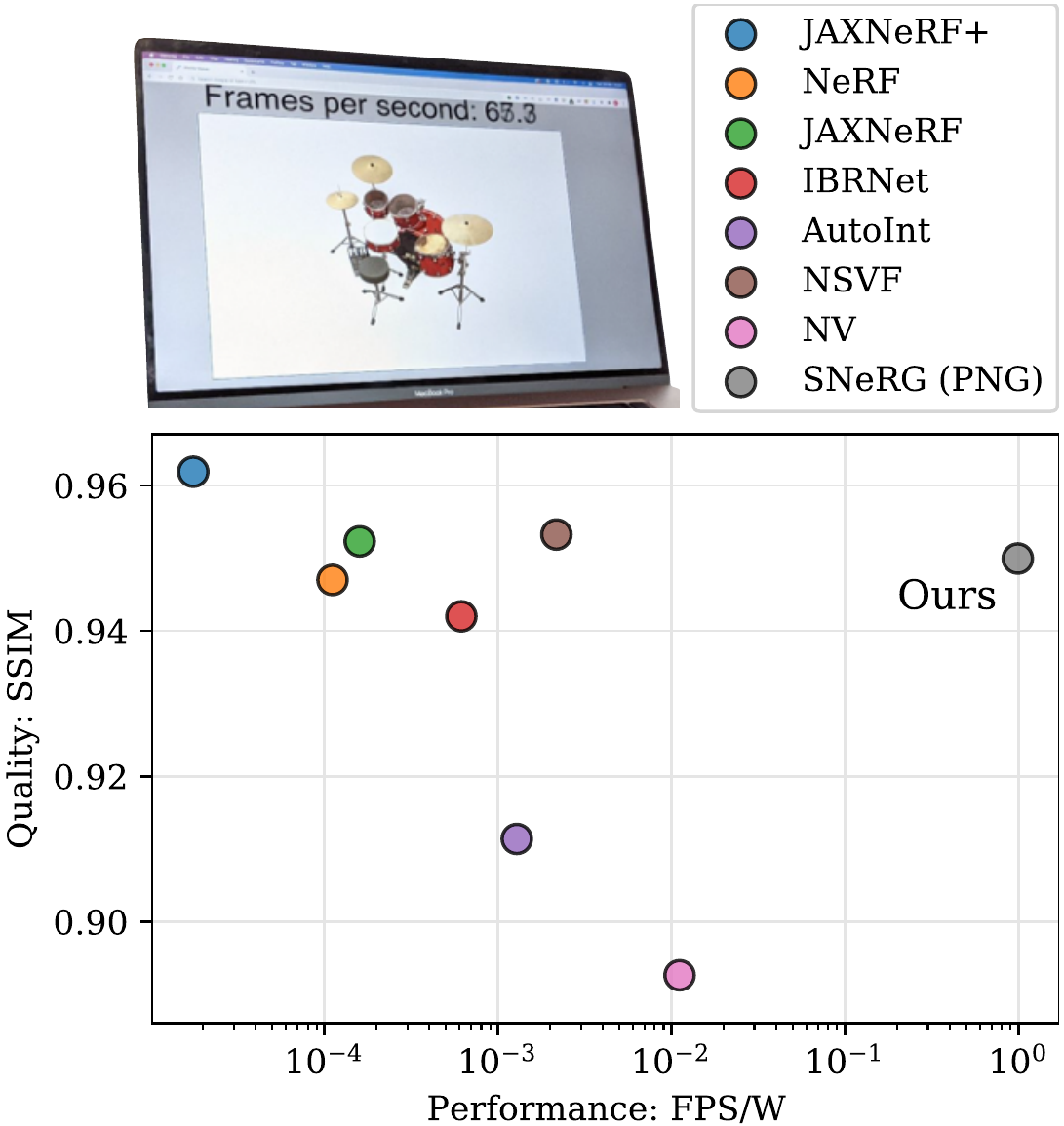}%
    \end{minipage}%
    \caption{Our method ``bakes'' NeRF's continuous neural volumetric scene representation into a discrete Sparse Neural Radiance Grid (SNeRG) for real-time rendering on commodity hardware ($\sim65$ frames per second on a MacBook Pro in the example shown here and in our supplemental video). Our method is more than two orders of magnitude faster than prior work for accelerating NeRF's rendering procedure and more than an order of magnitude faster than the next-fastest alternative (Neural Volumes) while producing substantially higher-quality renderings.}
    \label{fig:teaser}%
\end{figure}

In this paper, we address the problem of rendering a trained NeRF in real-time, see Figure~\ref{fig:teaser}. Our approach accelerates NeRF's rendering procedure by three orders of magnitude, resulting in a rendering time of 12 milliseconds per frame on a single GPU. We precompute and store (i.e. ``bake'') a trained NeRF into a sparse 3D voxel grid data structure, which we call a Sparse Neural Radiance Grid (SNeRG). Each active voxel in a SNeRG contains opacity, diffuse color, and a learned feature vector that encodes view-dependent effects. To render this representation, we first accumulate the diffuse colors and feature vectors along each ray. Next, we pass the accumulated feature vector through a lightweight MLP to produce a view-dependent residual that is added to the accumulated diffuse color.

We introduce two key modifications to NeRF that enable it to be effectively baked into this sparse voxel representation: 1) we design a ``deferred'' NeRF architecture that represents view-dependent effects with an MLP that only runs once per \emph{pixel} (instead of once per 3D sample as in the original NeRF architecture), and 2) we regularize NeRF's predicted opacity field during training to encourage sparsity, which improves both the storage cost and rendering time for the resulting SNeRG. 

We demonstrate that our approach is able to increase the rendering speed of NeRF so that frames can be rendered in real-time, while retaining NeRF's ability to represent fine geometric details and convincing view-dependent effects. Furthermore, our representation is compact, and requires less than 90 MB on average to represent a scene.


\section{Related work}

Our work draws upon ideas from computer graphics to enable the real-time rendering of NeRFs. In this section, we review scene representations used for view synthesis with a specific focus on their ability to support real-time rendering, and discuss prior work in efficient representation and rendering of volumetric representations within the field of computer graphics. 

\shortpara{Scene Representations for View Synthesis}
The task of view synthesis, using observed images of an object or scene to render photorealistic images from novel unobserved viewpoints, has a rich history within the fields of graphics and computer vision. The majority of prior work in this space has used traditional 3D representations from computer graphics which are naturally amenable to efficient rendering. For scenarios where the scene is captured by densely-sampled images, light field rendering techniques~\cite{davis12,gortler96,levoy96} can be used to efficiently render novel views by interpolating between sampled rays. Unfortunately, the sampling and storage requirements of light field interpolation techniques are typically intractable for settings with significant viewpoint motion. Methods that aim to support free-viewpoint rendering from sparsely-sampled images typically reconstruct an explicit 3D representation of the scene~\cite{debevec96}. One popular class of view synthesis methods uses mesh-based representations, with either diffuse~\cite{waechter2014} or view-dependent~\cite{buehler01, debevec96, wood00} appearance. Recent methods have trained deep networks to increase the quality of mesh renderings, improving robustness to errors in the reconstructed mesh geometry~\cite{hedman2018,thies2019neural}. Mesh-based approaches are naturally amenable to real-time rendering with highly-optimized rasterization pipelines. However, gradient-based optimization of a rendering loss with a mesh representation is difficult, so these methods have difficulties reconstructing fine structures and detailed scene geometry.

Another popular class of view synthesis methods uses discretized volumetric representations such as voxel grids~\cite{lombardi19, seitz99, sitzmann19, srinivasan20} or multiplane images~\cite{flynn19,penner17,srinivasan19,zhou18}. While volumetric approaches are better suited to gradient-based optimization, discretized voxel representations are fundamentally limited by their cubic scaling. This restricts their usage to representing scenes at relatively low resolutions in the case of voxel grids, or rendering from a limited range of viewpoints in the case of multiplane images. 

NeRF~\cite{mildenhall20} proposes replacing these discretized volumetric representations with an MLP that represents a scene as a \emph{continuous neural volumetric function} by mapping from a 3D coordinate to the volume density and view-dependent emitted radiance at that position. The NeRF representation has been remarkably successful for view synthesis, and follow-on works have extended NeRF for generative modeling~\cite{chan2020pi, schwarz2020graf}, dynamic scenes~\cite{li2021, ost2020neural}, non-rigidly deforming objects~\cite{gafni2020dynamic, park2020deformable}, and relighting~\cite{bi2020, nerv2021}. NeRF is able to represent detailed geometry and realistic appearance extremely efficiently (NeRF uses approximately 5 MB of network weights to represent each scene), but this comes at the cost of slow rendering. NeRF needs to query its MLP hundreds of times per ray, and requires roughly a minute to render a single frame. We specifically address this issue, and present a method that enables a trained NeRF to be rendered in real-time. 

Recent works have explored a few strategies for improving the efficiency of NeRF's neural volumetric rendering. AutoInt~\cite{lindell2021autoint} designs a network architecture that automatically computes integrals along rays, which enables a piecewise ray-marching procedure that requires far fewer MLP evaluations. Neural Sparse Voxel Fields~\cite{liu2020nsvf} store a 3D voxel grid of latent codes, and sparsifies this grid during training to enable NeRF to skip free space during rendering. Decomposed Radiance Fields~\cite{rebain21} represents a scene using a set of smaller MLPs instead of a single large MLP. However, these methods only achieve moderate speedups of around $10\times$ at best, and are therefore not suited for real-time rendering. In contrast to these methods, we specifically focus on accelerating the rendering of a NeRF after it has been trained, which allows us to leverage precomputation strategies that are difficult to incorporate during training.

\begin{figure*}[t]
    \centering
    \includegraphics[width=\linewidth]{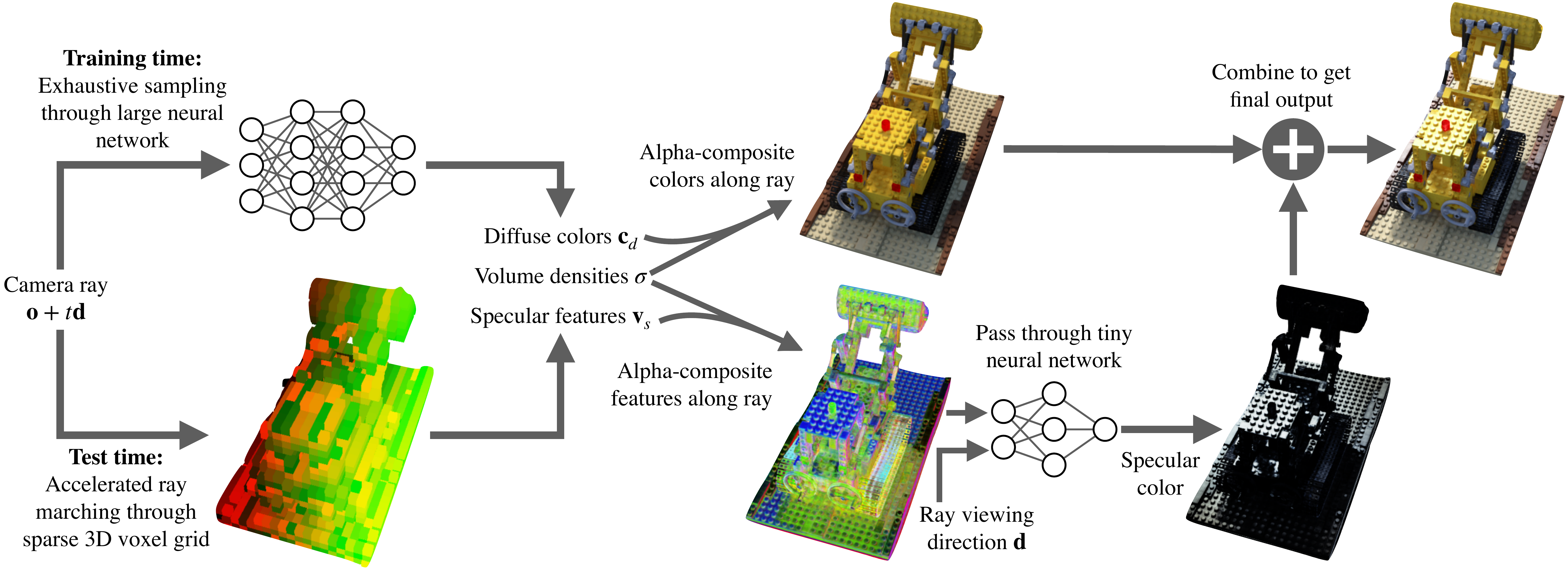}
    \caption{
    \textbf{Rendering pipeline overview.} We jointly design a ``deferred'' NeRF as well as a procedure to precompute and store the outputs of a trained deferred NeRF in a sparse 3D voxel grid data structure for real-time rendering. At training time, we query the deferred NeRF's MLP for the diffuse color, volume density, and feature-vector at any 3D location along a ray. At test time, we instead precompute and store these values in a sparse 3D voxel grid data structure. Next, we alpha-composite the diffuse colors and feature vectors along the ray. Once the ray has terminated, we use another MLP to predict a view-dependent specular color from the accumulated feature vector, diffuse color, and the ray's viewing direction. Note that this MLP is only run once per pixel.}
    \label{fig:render_pipeline}
\end{figure*}

\shortpara{Efficient Volume Rendering}
Discretized volumetric representations have been used extensively in computer graphics to make rendering more efficient both in terms of storage and rendering speed. Our representation is inspired by this long line of prior work on efficient volume rendering, and we extend these approaches with a deferred neural rendering technique to model view-dependent effects.

Early works in volume rendering~\cite{lacroute94, levoy90, Westover1991} primarily focused on fast rendering of dense voxel grids. However, as shown by Laine and Karras~\cite{laine10}, sparse voxel grids can be an effective and efficient representation for opaque surfaces. In situations where large regions of space share the same data value or a prefiltered representation is required to combat aliasing, hierarchical representations such as sparse voxel octrees~\cite{Crassin2009} are a popular choice of data structure to represent this sparse volumetric content. However, for scenes with detailed geometry and appearance and scenarios where a variable level-of-detail is not required during rendering, octrees' intermediate non-leaf nodes and the tree traversals required to query them can incur a significant memory and time overhead.

Alternatively, sparse voxel grids can be efficiently represented with hash tables~\cite{lefebvre06, niessner13}. However, hashing each voxel independently can lead to incoherent memory fetches when traversing the representation during rendering. 
We make a deliberate trade-off to use a \emph{block-sparse} representation, which improves memory coherence but slightly increases the size of our representation.

Our work aims to combine the reconstruction quality and view-dependence of NeRF with the speed of these efficient volume rendering techniques. We achieve this by extending deferred neural rendering~\cite{thies2019neural} to volumetric scene representations. This allows us to visualize trained NeRF models in real-time on commodity hardware, with minimal quality degradation. 


\section{Method Overview}

Our overall goal is to design a practical representation that enables the serving and real-time rendering of scenes reconstructed by NeRF. This implies three requirements: 1) Rendering a $800\times800$ resolution frame (the resolution used by NeRF) should require less than 30 milliseconds on commodity hardware. 2) The representation should be compressible to 100 MB or less. 3) The uncompressed representation should fit within GPU memory (approximately 4 GB) and should not require streaming.

Rendering a standard NeRF in real-time is completely intractable on current hardware. NeRF requires about 100 teraflops to render a single $800\times800$ frame, which results in a best-case rendering time of 10 seconds per frame on an NVIDIA RTX 2080 GPU with full GPU utilization. To enable real-time rendering, we must therefore exchange some of this computation for storage. However, we do not want to precompute and store the entire 5D view-dependent representation~\cite{gortler96, levoy96}, as that would require a prohibitive amount of GPU memory.

We propose a hybrid approach that precomputes and stores some content in a sparse 3D data structure but defers the computation of view-dependent effects to rendering time. We jointly design a reformulation of NeRF (Section~\ref{sec:nerf}) as well as a procedure to bake this modified NeRF into a discrete volumetric representation that is suited for real-time rendering (Section~\ref{sec:snerg}).

\section{Modifying NeRF for Real-time Rendering}
\label{sec:nerf}
We reformulate NeRF in three ways: 1) we limit the computation of view-dependent effects to a single network evaluation \emph{per ray}, 2) we introduce a small bottleneck in the network architecture that can be efficiently stored as 8 bit integers, and 3) we introduce a sparsity loss during training, which concentrates the opacity field around surfaces in the scene. Here, we first review NeRF's architecture and rendering procedure before describing our modifications. 

\subsection{Review of NeRF}

NeRF represents a scene as a continuous volumetric function parameterized by a MLP. Concretely, the 3D position $\ray(t)$ and viewing direction $\raydir$ along a camera ray $\ray(t) = \rayorigin + t \raydir$, are passed as inputs to an MLP with weights $\modelweights$ to produce the volume density $\sigma$ of particles at that location as well as the RGB color $\col$ corresponding to the radiance emitted by particles at the input location along the input viewing direction:
\begin{equation}
    \sigma(t), \col(t) = \mlp_\modelweights\left( \ray(t),\raydir \right)\, .
\end{equation}
A key design decision made in NeRF is to architect the MLP such that volume density is only predicted as a function of 3D position, while emitted radiance is predicted as a function of both 3D position and 2D viewing direction.

To render the color $\hat{\Col}(\ray)$ of a pixel, NeRF queries the MLP at sampled positions $t_k$ along the corresponding ray and uses the estimated volume densities and colors to approximate a volume rendering integral using numerical quadrature, as discussed by Max~\cite{max95}:
\begin{gather}
    \label{eqn:nerfrender}
    \hat{\Col}(\ray) = \sum_{k} T(t_k) \, \decay{\sigma(t_k) \delta_{k}} \, \col(t_k) \, ,\\
    T(t_k) = \exp \left(\!-\!\sum_{k'=1}^{k-1} \sigma(t_{k'}) \delta_{k'}\!\right) \,, \quad \decay{x} = 1-\exp(-x)\,, \nonumber
\end{gather}
where $\delta_k = t_{k+1} - t_k$ is the distance between two adjacent points along the ray.

NeRF trains the MLP by minimizing the squared error between input pixels from a set of observed images (with known camera poses) and the pixel values predicted by rendering the scene as described above:
\begin{equation}
    \mathcal{L}_r = \sum_{i} \normsq{\truecol(\ray_{i}) -  \hat{\Col}(\ray_{i})}
\end{equation}
where $\truecol(\ray_{i})$ is the color of pixel $i$ in the input images.

By replacing a traditional discrete volumetric representation with an MLP, NeRF makes a strong space-time tradeoff: NeRF's MLP requires multiple orders of magnitude less space than a dense voxel grid, but accessing the properties of the volumetric scene representation at any location requires an MLP evaluation instead of a simple memory lookup. Rendering a single ray that passes through the volume requires hundreds of these MLP queries, resulting in extremely slow rendering times. This tradeoff is beneficial during training; since we do not know where the scene geometry lies during optimization, it is crucial to use a compact representation that can represent highly-detailed geometry at arbitrary locations. However, after a NeRF has been trained, we argue that it is prudent to rethink this space-time tradeoff and bake the NeRF representation into a data structure that stores pre-computed values from the MLP to enable real-time rendering.

\subsection{Deferred NeRF Architecture}

NeRF's MLP can be thought of as predicting a 256-dimensional feature vector for each input 3D location, which is then concatenated with the viewing direction and decoded into an RGB color. NeRF then accumulates these view-dependent colors into a single pixel color. However, evaluating an MLP at every sample along a ray to estimate the view-dependent color is prohibitively expensive for real-time rendering. Instead, we modify NeRF to use a strategy similar to deferred rendering~\cite{Deering:1988:TPN,thies2019neural}. We restructure NeRF to output a diffuse RGB color $\col_d$ and a 4-dimensional feature vector $\feat$ (which is constrained to $[0,1]$ via a sigmoid so that it can be compressed, as discussed in Section~\ref{sec:compression}) in addition to the volume density $\sigma$ at each input 3D location:
\begin{equation}
    \sigma(t), \col_d(t), \feat(t) = \mlp_\modelweights\left( \ray(t) \right)\, .
\end{equation}
To render a pixel, we accumulate the diffuse colors and feature vectors along each ray and pass the accumulated feature vector and color, concatenated to the ray's direction, to a very small MLP with parameters $\modelweightsspec$ (2 layers with 16 channels each) to produce a view-dependent residual that we add to the accumulated diffuse color:
\begin{gather}
    \hat{\Col}_{d}(\ray) = \sum_{k} T(t_k) \, \decay{\sigma(t_k) \delta_{k}} \, \col_d(t_k) \, ,\\
    \Feat(\ray) = \sum_{k} T(t_k) \, \decay{\sigma(t_k) \delta_{k}} \, \feat(t_k) \, ,\\
    \label{eqn:defrender}
     \hat{\Col}(\ray) =  \hat{\Col}_{d}(\ray) + \mlp_\modelweightsspec \left( \Feat(\ray), \raydir \right)
\end{gather}
This modification enables us to precompute and store the diffuse colors and 4-dimensional feature vectors within our sparse voxel grid representation discussed below. Critically, we only need to evaluate the $\mlp_\modelweightsspec$ to produce view-dependent effects once per \emph{pixel}, instead of once per sample in 3D space as in the standard NeRF model. 

\subsection{Opacity Regularization}

Both the rendering time and required storage for a volumetric representation strongly depend on the sparsity of opacity within the scene. To encourage NeRF's opacity field to be sparse, we add a regularizer that penalizes predicted density using a Cauchy loss during training:
\begin{equation}
\mathcal{L}_s = \lambda_s \sum_{i, k} \text{log}\left(1 + \frac{\sigma(\ray_i (t_k))^2}{c} \right) \, ,
\end{equation}
where $i$ indexes pixels in the input (training) images, $k$ indexes samples along the corresponding rays, and hyperparameters $\lambda_s$ and $c$ control the magnitude and scale of the regularizer respectively ($\lambda_s = 10^{-4}$ and $c=\nicefrac{1}{2}$ in all experiments). To ensure that this loss is not unevenly applied due to NeRF's hierarchical sampling procedure, we only compute it for the ``coarse'' samples that are distributed with uniform density along each ray.

\begin{figure}[t]
    \centering
    \includegraphics[width=\linewidth]{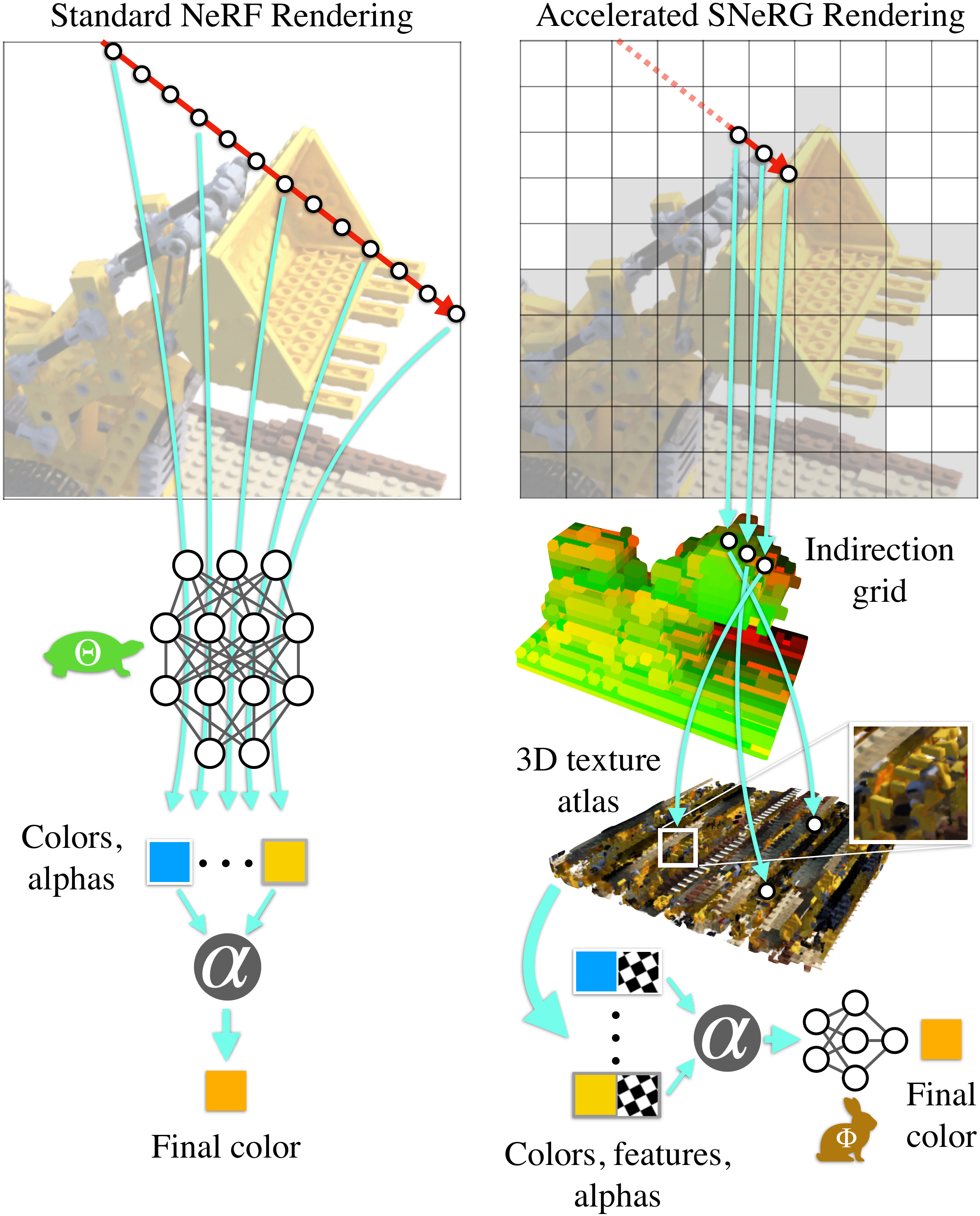}
    \caption{
    \textbf{Comparison of ray-marching procedures for NeRF and SNeRG.} Left: To render a ray with NeRF, we densely sample points along the ray and pass the coordinates to a large MLP to compute colors and opacities, which are alpha-composited into a pixel color. Right: SNeRG significantly accelerates rendering by replacing compute-intensive MLP evaluations with lookups into a precomputed sparse 3D grid representation. We use an indirection grid to map occupied voxel blocks to locations within a compact 3D texture atlas. During ray-marching, we skip unoccupied blocks, and alpha-composite the diffuse colors and feature vectors fetched along the ray. We terminate ray-marching once the accumulated opacity saturates, and pass the accumulated color and features to a small MLP that evaluates the view-dependent component for the ray.}
    \label{fig:marching_pancake}
\end{figure}

\section{Sparse Neural Radiance Grids}
\label{sec:snerg}

We now convert a trained deferred NeRF model, described above, into a representation suitable for real-time rendering. The core idea is to trade computation for storage, significantly reducing the time required to render frames. In other words, we are looking to replace the MLP evaluations in NeRF with fast lookups in a precomputed data structure.
We achieve this by precomputing and storing, i.e. \emph{baking}, the diffuse colors $\col_d$, volume densities $\sigma$, and 4-dimensional feature vectors $\feat$ in a voxel grid data structure. 

It is crucial for us to store this volumetric grid using a sparse representation, as a dense voxel grid can easily fill up all available memory on a modern high-end GPU. By exploiting sparsity and only storing voxels that are both occupied and visible, we end up with a much more compact representation.

\subsection{SNeRG Data Structure}

\newcommand{\hotdogpic}[1]{\includegraphics[width=1.6in]{figures/sparsity/#1_joncrop.png}}

\begin{figure}
    \centering
    \begin{tabular}{@{}c@{}c@{}}
    \hotdogpic{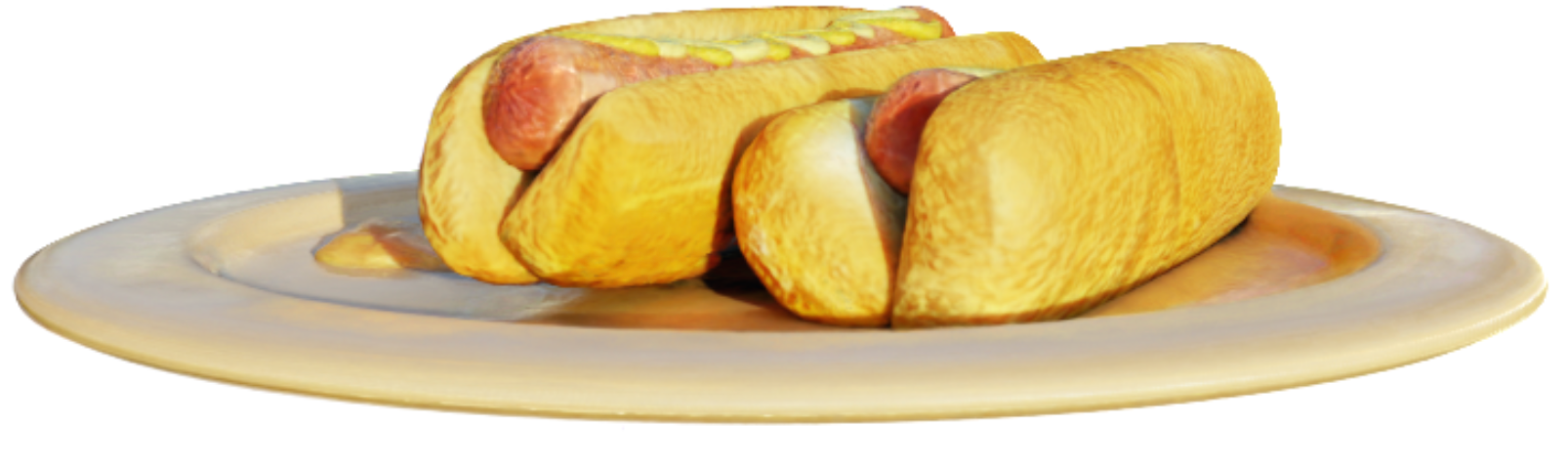} & \hotdogpic{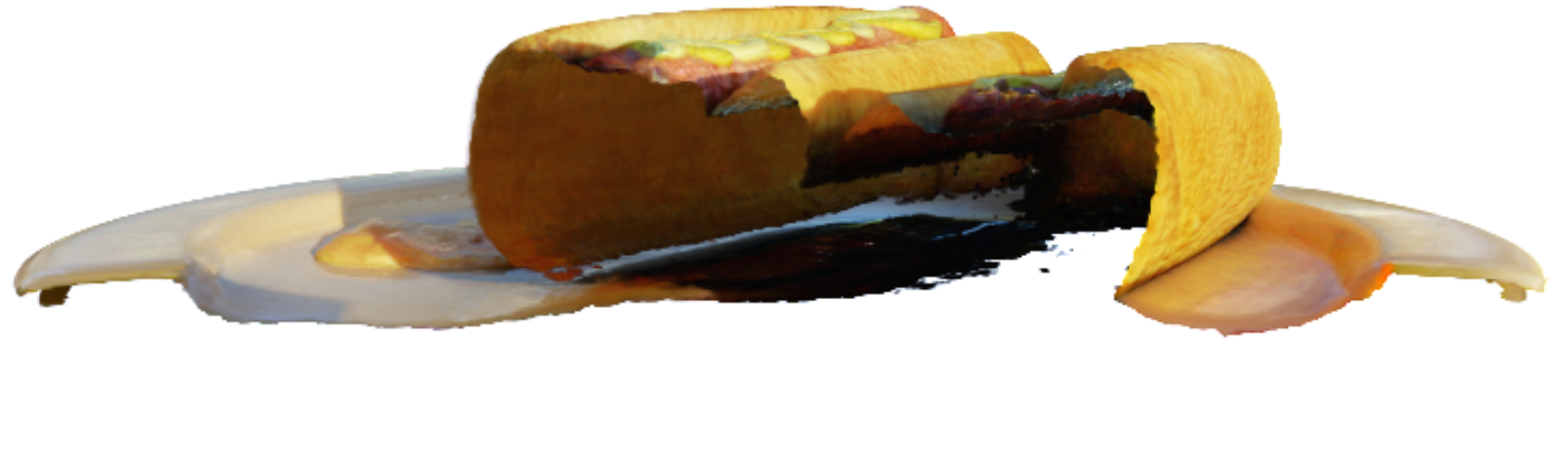} \\
    {\footnotesize (a) Frame rendered by our method.} & \footnotesize{(b) Cross-section  of (a).} \\
    \hotdogpic{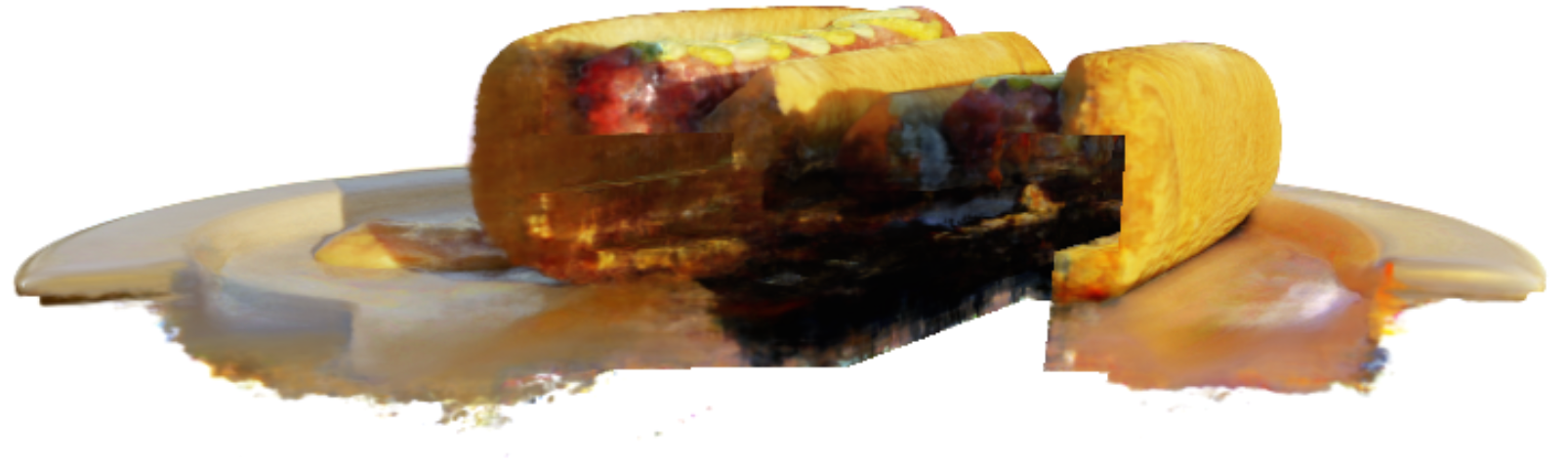} & \hotdogpic{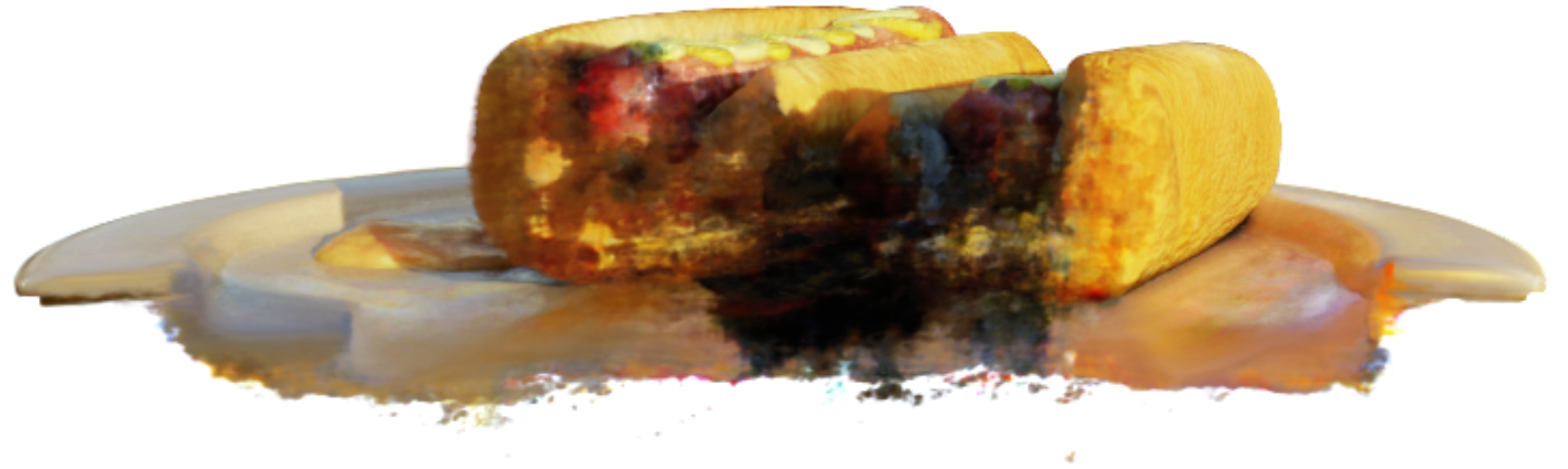} \\
    {\footnotesize (c) Trained without $\mathcal{L}_s$.} & 
    {\footnotesize (d) No $\mathcal{L}_s$, no visibility culling.}
    \end{tabular}
    \caption{\textbf{Visualization of sparsity loss and visibility culling.} We render cross-sections of the \scenename{Hotdog} scene to inspect the effect of our sparsity loss $\mathcal{L}_s$ and visibility culling. Our full method (b) represents the scene by only allocating content around visible scene surfaces. Removing either the sparsity loss alone (c) or both the sparsity loss and visibility culling (d) results in a much less compact representation.}
    \label{fig:sparsity}
\end{figure}

Our Sparse Neural Radiance Grid (SNeRG) data structure represents an $N^3$ voxel grid in a block-sparse format using two smaller dense arrays.

The first array is a 3D texture atlas containing densely-packed ``macroblocks'' of size $B^3$ each, corresponding to the content (diffuse color, feature vectors, and opacity) that actually exists in the sparse volume. Each voxel in the 3D atlas represents the scene at the full resolution of the dense $N^3$ grid, but the 3D texture atlas is much smaller than $N^3$ since it only contains the sparse ``occupied'' content. Compared to hashing-based data structures (where $B^3=1$), this approach helps keep spatially close content nearby in memory, which is beneficial for efficient rendering.

The second array is a low resolution $(\nicefrac{N}{B})^3$ indirection grid, which either stores a value indicating that the corresponding $B^3$ macroblock within the full voxel grid is empty, or stores an index that points to the high-resolution content of that macroblock within the 3D texture atlas. This structure crucially lets us skip blocks of empty space during rendering, as we describe below.

\subsection{Rendering}
\label{sec:grid_rendering}

We render a SNeRG using a ray-marching procedure, as done in NeRF. The critical differences that enable real-time rendering are: 1) we precompute the diffuse colors and feature vectors at each 3D location, allowing us to look them up within our data structure instead of evaluating an MLP, and 2) we only evaluate an MLP to produce view-dependent effects once per pixel, as opposed to once per 3D location.

To estimate the color of each ray, we first march the ray through the indirection grid, skipping macroblocks that are marked as empty. For macroblocks that are occupied, we step at the voxel width $\nicefrac{1}{N}$ through the corresponding block in the 3D texture atlas, and use trilinear interpolation to fetch values at each sample location. We further accelerate rendering and conserve memory bandwidth by only fetching features where the volume density is non-zero.
We use standard alpha compositing to accumulate the diffuse color and features, terminating ray-marching once the opacity has saturated. Finally, we compute the view-dependent specular color for the ray by evaluating $\mlp_\modelweightsspec$ with the accumulated color, feature vector and the ray's viewing direction. We then add the resulting residual color to the accumulated diffuse color, as described in Equation~\ref{eqn:defrender}.

\subsection{Baking}
\label{sec:baking}

To minimize storage cost and rendering time, our baking procedure aims to only allocate storage for voxels in the scene that are both non-empty and visible in at least one of the training views. 
We start by densely evaluating the NeRF network for the full $N^3$ voxel grid. We convert NeRF's unbounded volume density values, $\sigma$, to traditional opacity values $\alpha=1-\exp(\sigma v)$, where $v=\nicefrac{1}{N}$ is the width of a voxel. 
Next, we sparsify this voxel grid by culling empty space, i.e. macroblocks where the maximum opacity is low (below $\tau_\alpha$), and culling macroblocks for which the voxel visibilities are low (maximum transmittance $T$ between the voxel and all training views is below $\tau_{vis}$). In all experiments, we set $\tau_\alpha=0.005$ and $\tau_{vis} = 0.01$.
Finally, we compute an anti-aliased estimate for the content in the remaining macroblocks by densely evaluating the trained NeRF at 16 Gaussian distributed locations within each voxel ($\sigma=\nicefrac{v}{\sqrt{12}}$) and averaging the resulting diffuse colors, feature vectors, and volume densities.

\subsection{Compression}
\label{sec:compression}

We quantize all values in the baked SNeRG representation to 8 bits and separately compress the indirection grid and the 3D texture atlas. We compress each slice of the indirection grid as a lossless PNG, and we compress the 3D texture atlas as either a set of lossless PNGs, a set of JPEGs, or as a single video encoded with H264. The quality versus storage tradeoff of this choice is evaluated in Table~\ref{tbl:storage_ablation}. For synthetic scenes, compressing the texture atlas results in approximately $80\times$, $100\times$, and $230\times$ compression rates for PNG, JPEG, and H264, respectively. We specifically choose a macroblock size of $32^3$ voxels to align the 3D texture atlas macroblocks with the blocks used in image compression. This reduces the size of the compressed 3D texture atlas because additional coefficients are not needed to represent discontinuities between macroblocks.

\begin{figure}
\captionsetup[sub]{font=small}
     \centering
     \begin{subfigure}[b]{0.32\columnwidth}
         \centering
         \includegraphics[width=\textwidth]{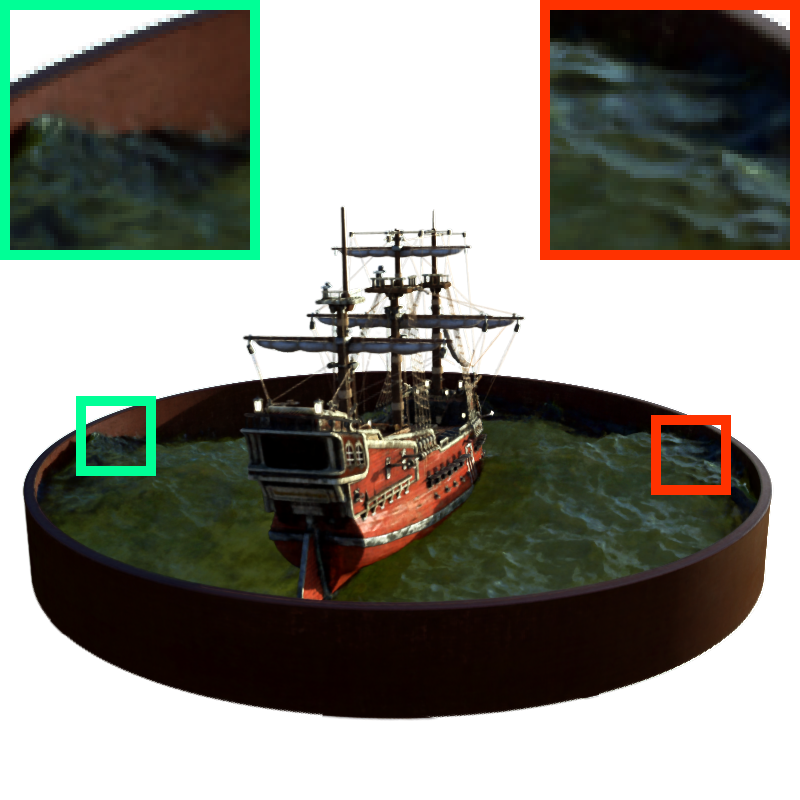}
         \caption{No FT (23.30)}
     \end{subfigure}
     \begin{subfigure}[b]{0.32\columnwidth}
         \centering
         \includegraphics[width=\textwidth]{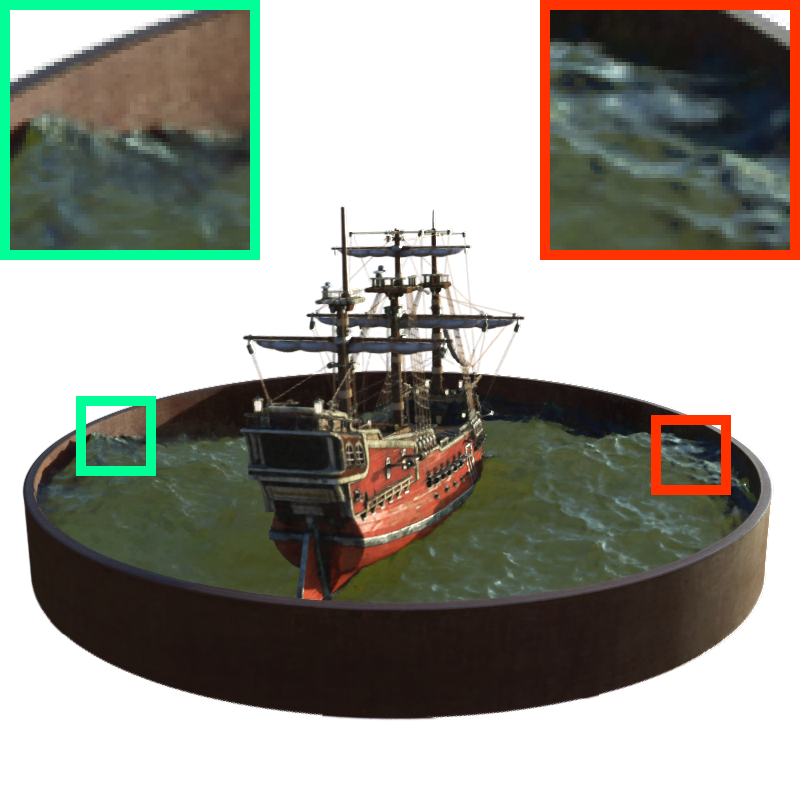}
         \caption{FT (28.35)}
     \end{subfigure}
     \begin{subfigure}[b]{0.32\columnwidth}
         \centering
         \includegraphics[width=\textwidth]{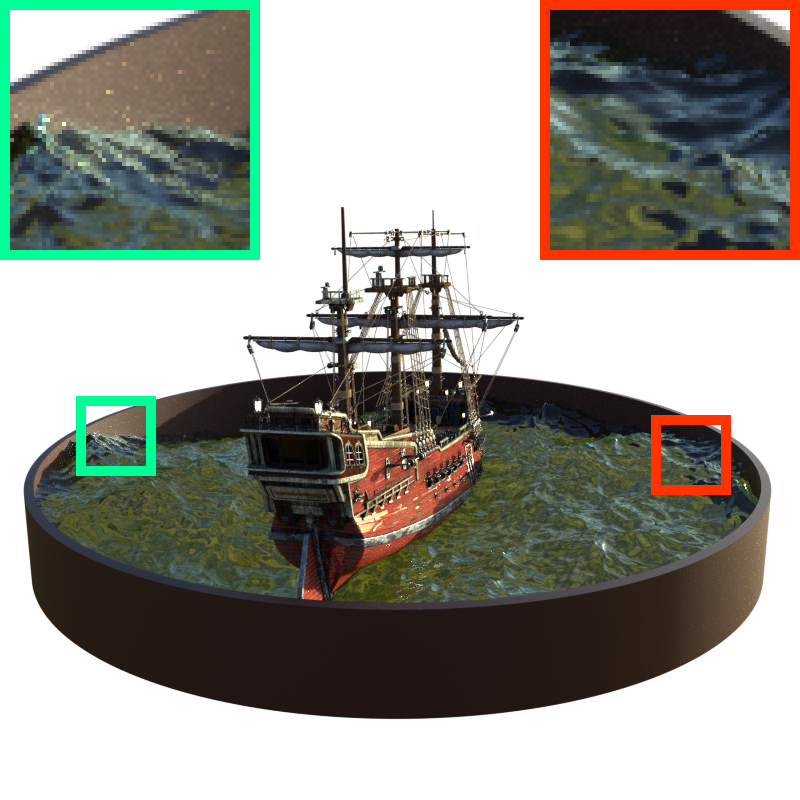}
         \caption{Ground Truth}
     \end{subfigure}
        \caption{\textbf{Impact of fine-tuning (FT) the view-dependent appearance network} (PSNR in parentheses). (a) Renderings from a SNeRG representation can be lower-quality than those from the deferred NeRF, primarily due the quantization from 32-bit floating point to 8-bit integer values (see Sections~\ref{sec:baking}, \ref{sec:compression}). (b) We are able to regain most of the lost quality by fine-tuning the weights of the deferred shading network $\mlp_\modelweightsspec$ (Section~\ref{sec:refinement}).}
        \label{fig:refinement}
\end{figure}

\subsection{Fine-tuning}
\label{sec:refinement}

While the compression and quantization procedure described above is crucial for making SNeRG compact and easy to distribute, the quality of images rendered from the baked SNeRG is lower than the quality of images rendered from the corresponding deferred NeRF. Figure~\ref{fig:refinement} visualizes how quantization affects view-dependent effects by biasing renderings towards a darker, diffuse-only color $\hat{\Col}_{d}(\ray)$.

Fortunately, we are able to recoup almost all of that lost accuracy by fine-tuning the weights of the deferred per-pixel shading $\mlp_\modelweightsspec$ to improve the final rendering quality (Table~\ref{tbl:quality_ablation}). We optimize the parameters $\modelweightsspec$ to minimize the squared error between the observed input images used to train the deferred NeRF and the images rendered from our SNeRG. We use the Adam optimizer~\cite{adam} with a learning rate of $3 \times 10^{-4}$ and optimize for 100 epochs.

\begin{figure}
     \centering
     \includegraphics[width=\columnwidth]{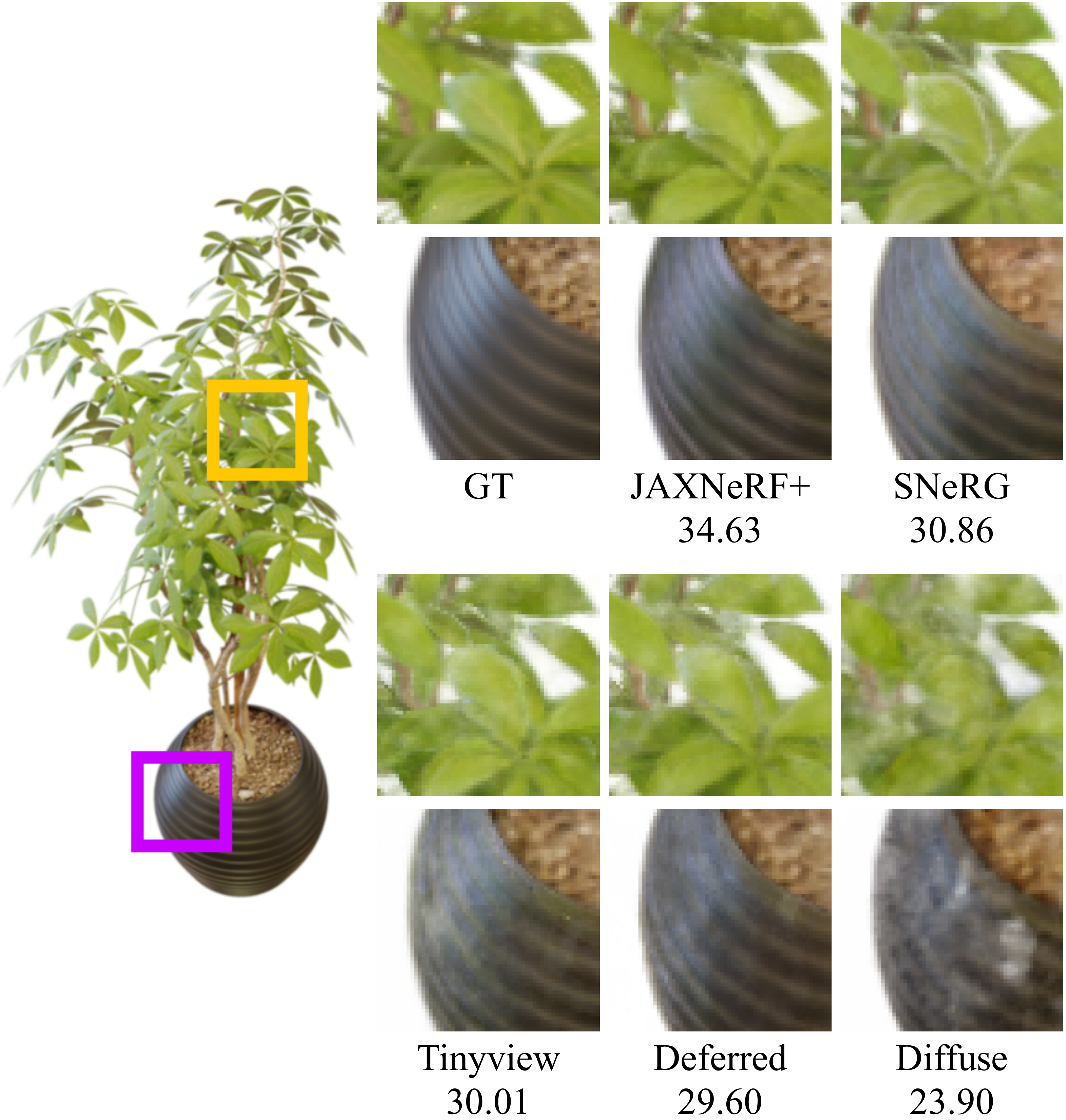}
        \caption{
        \textbf{Ablation study} showing visual examples from  Table~\ref{tbl:quality_ablation}. Note the minimal difference in quality between the full JAXNeRF+ network and the successive approximations we make to speed up view-dependence (\emph{Tinyview}, \emph{Deferred NeRF} and \emph{SNeRG}). For completeness, we show the floating alpha artifacts introduced by not modelling view-dependence at all (\emph{Diffuse}).
        }
        \label{fig:ablation}
\end{figure}

\section{Implementation Details}

Our deferred NeRF model is based on JAXNeRF~\cite{jaxnerf2020github}, an implementation of NeRF in JAX~\cite{jax2018github}. As in NeRF, we apply a positional encoding~\cite{tancik20} to positions and view directions. We train all networks for 250k iterations with a learning rate which decays log-linearly from $2\times10^{-3}$ to $2\times10^{-5}$. To improve stability we use JAXNeRF's ``warm up'' functionality to reduce the learning rate to $2\times10^{-4}$ for the first 2500 iterations, and we clip gradients by value (at $0.01$) and then by norm (also at $0.01$). We use a batch size of 8,192 for synthetic scenes and a batch size of 16,384 for real scenes.

As our rendering time is independent of $\mlp_\modelweights$'s model size, we can afford to use a larger network for our experiments. To this end, we base our method on the JAXNeRF+ model, which was trained with 576 samples per ray (192 coarse, 384 fine) and uses 512 channels per layer in $\mlp_\modelweights$.

During baking, we set the voxel grid resolution to be slightly larger than the size of the training images: $1000^3$ for the synthetic scenes, and $1300^3$ for the real datasets. We implement the SNeRG renderer in Javascript and WebGL using the THREE.js library. We load the indirection grid and 3D texture atlas into 8 bit 3D textures. The view-dependence $\mlp_\modelweightsspec$ is stored uncompressed and is implemented in a WebGL shader. In all experiments, we run this renderer on a 2019 MacBook Pro laptop equipped with a 85W AMD Radeon Pro 5500M GPU.

\begin{figure}
     \centering
\begin{tabular}{@{}c@{\,}c@{}}
    \includegraphics[width=0.49\columnwidth]{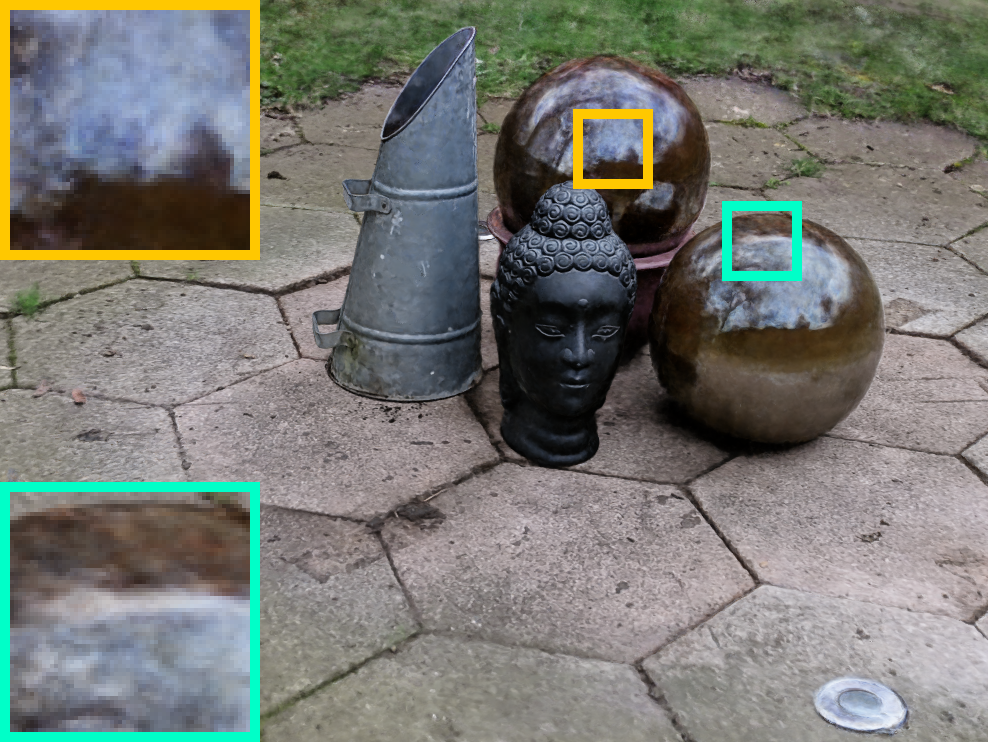} & 
    \includegraphics[width=0.49\columnwidth]{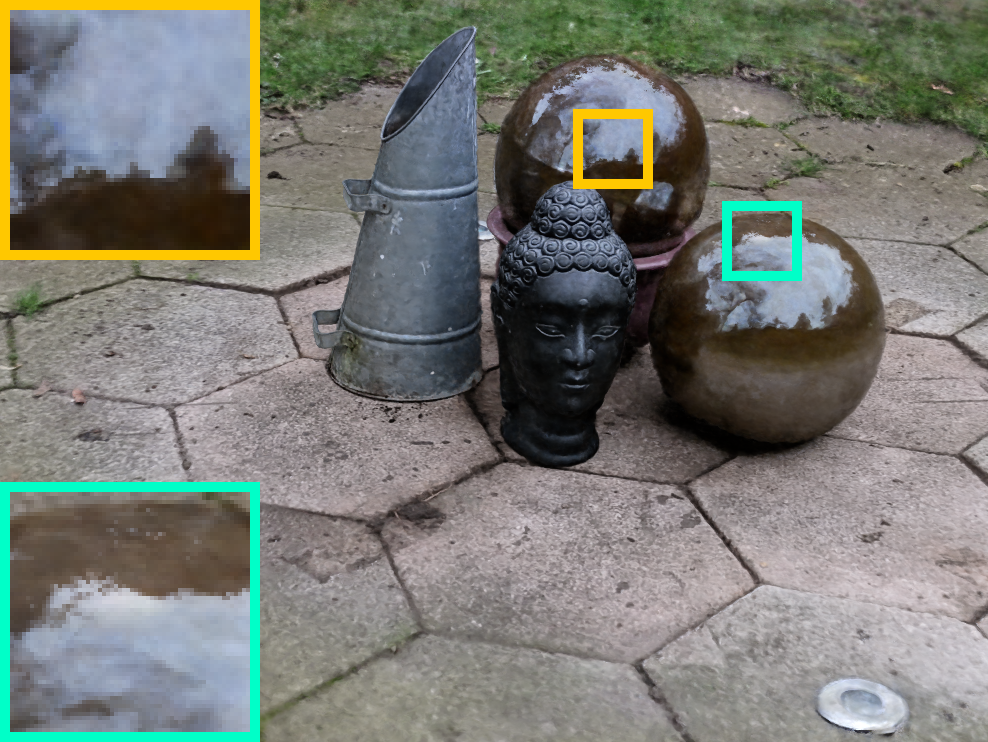} \\
    \small (a) JAXNeRF+ (23.00) & 
    \small (b) Deferred (22.75) \\ 
    \includegraphics[width=0.49\columnwidth]{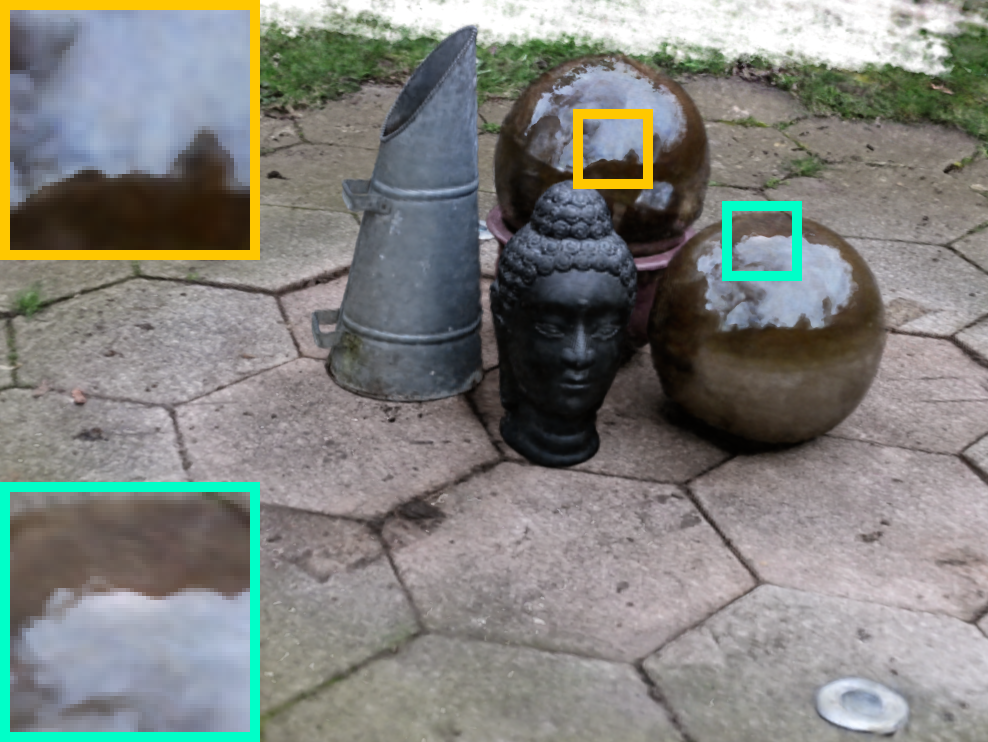} & 
    \includegraphics[width=0.49\columnwidth]{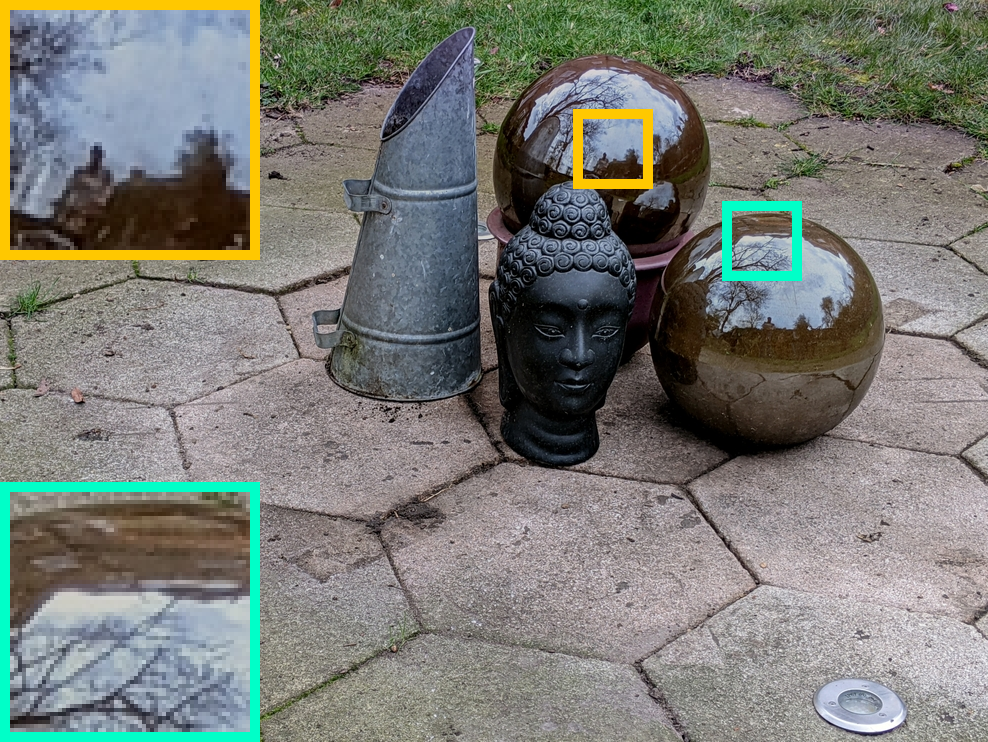} \\
    \small (c) SNeRG (21.45) & 
    \small (d) Ground Truth \\ 
    \end{tabular}
    \vspace{-0.1in}
        \caption{\textbf{Real $\mathbf{360^\circ}$ scene} (PSNR in parentheses). Note how our real-time method is able to model the mirror-like reflective surface of the garden spheres.}
        \label{fig:real360}
\end{figure}

\section{Experiments}

We validate our design decisions with an extensive set of ablation studies and comparisons to recent techniques for accelerating NeRF. Our experiments primarily focus on free-viewpoint rendering of $360^\circ$ scenes (scenes captured by inwards-facing cameras on the upper hemisphere).
Though acceleration techniques already exist for the special case in which all cameras face the same direction (see Broxton~\etal~\cite{broxton20}), $360^\circ$ scenes represent a challenging and general use-case that has not yet been addressed. In Figure~\ref{fig:real360}, we show an example of a real $360^{\circ}$ scene and present more results in our supplement, including the forward-facing scenes from Local Light Field Fusion (LLFF)~\cite{mildenhall19}.

We evaluate all ablations and baseline methods according to three criteria: render-time performance (measured by frames per second as well as GPU memory consumption in gigabytes), storage cost (measured by megabytes required to store the compressed representation), and rendering quality (measured using the PSNR, SSIM~\cite{wang2004image}, and LPIPS~\cite{zhang2018unreasonable} image quality metrics).
It is important to explicitly account for power consumption when evaluating performance --- algorithms that are fast on a high performance GPU are not necessarily fast on a laptop. We therefore adopt the convention used by the high performance graphics community of measuring performance relative to power consumption, i.e. FPS per watt, or equivalently, frames per joule~\cite{Johnsson2012}.

Please refer to our video for screen captures of our technique being used for real-time rendering on a laptop.




\subsection{Ablation Studies}

In Table~\ref{tbl:performance_ablation}, we ablate combinations of three components of our method that primarily affect speed and GPU memory usage. Ablation 1 shows that removing the view-dependence MLP has a minimal effect on runtime performance. Ablation 2 shows that removing the sparsity loss $\mathcal L_s$ greatly increases (uncompressed) memory usage. Ablation 3 shows that switching from our ``deferred'' rendering back to NeRF's approach of querying an MLP at each sample along the ray results in prohibitively large render times.

Table~\ref{tbl:quality_ablation} and Figure~\ref{fig:ablation} show the impact on rendering quality of each of our design decisions in building a representation suitable for real-time rendering. Although our simplifications of using a deferred rendering scheme (``Deferred'') and a smaller network architecture (``Tinyview'') for view-dependent appearance do slightly reduce rendering quality, they are crucial for enabling real-time rendering, as discussed above. Note that the initial impact on quality from quantizing and compressing our representation is significant. However, after fine tuning (``FT''), the final rendering quality of our SNeRG model remains competitive with the neural model from which it was derived (``Deferred'').

In Table~\ref{tbl:storage_ablation} we explore the impact of various compression schemes on disk storage space requirements. Our sparse voxel grid benefits greatly from applying compression techniques such as JPEG or H264 to its 3D texture atlas, achieving a file size over $200\times$ more compact than a naive 32 bit float array while sacrificing less than 1dB of PSNR. Because our sparsity loss $\mathcal L_s$ concentrates opaque voxels around surfaces (see Figure~\ref{fig:sparsity}), ablating it significantly increases model size. Our compressed SNeRG representations are small enough to be quickly loaded in a web page.

The positive impact of training our models using the sparsity loss $\mathcal{L}_s$ is visible across these ablations --- it more than doubles rendering speed, halves the storage requirements of both the compressed representation on disk and the uncompressed representation in GPU memory, and minimally impacts rendering quality.

\definecolor{LightCyan}{rgb}{0.88,1,1}
\begin{table}[]
\centering
\resizebox{\linewidth}{!}{
\begin{tabular}{r|ccc|r@{}lc}
       & MLP & $\mathcal{L}_s$ & Defer & \multicolumn{2}{c}{ms/frame $\downarrow$} & GPU GB  $\downarrow$ \\ \hline 
Ours &\cmark &\cmark &\cmark &  {11.9} &$\pm$  {4.5} & 1.73 $\pm$  1.48  \\ 
1) &\xmark &\cmark &\xmark &  {9.2} &$\pm$  {4.6} & 1.73 $\pm$  1.48 \\
2) & \cmark  &\xmark &\cmark &  {20.0}* &$\pm$  {5.3} & 4.26 $\pm$ 1.56  \\
3) & \cmark  &\cmark &\xmark &  343.6 &$\pm$  {247.5} & 1.73 $\pm$  1.48  \\
\end{tabular}
}
\caption{Performance ablation study, including uncompressed GPU memory used during rendering, on the Synthetic $360^\circ$ scenes. Ablation 2 ran out of memory on the \scenename{Ficus} scene, which biases the average runtime for that row.}
\label{tbl:performance_ablation}
\end{table}

\begin{table}[]
\centering
\small
\begin{tabular}{l|ccc}
        & PSNR $\uparrow$ & SSIM $\uparrow$ & LPIPS $\downarrow$ \\ \hline 
JAXNeRF+& \cellcolor{red} 33.00 & \cellcolor{red} 0.962 & \cellcolor{red} 0.038\\
JAXNeRF+ Tinyview& \cellcolor{orange} 31.65 & \cellcolor{orange} 0.954 & \cellcolor{orange} 0.047\\
JAXNeRF+ Deferred& \cellcolor{yellow} 30.55 & \cellcolor{yellow} 0.952 & \cellcolor{yellow} 0.049\\
SNeRG (PNG)& 30.38 & 0.950 & 0.050\\
SNeRG (PNG, no $\mathcal{L}_s$)& 30.22 & 0.949 & 0.050\\
SNeRG (PNG, no FT)& 26.68 & 0.930 & 0.053\\
JAXNeRF+ Diffuse& 27.39 & 0.927 & 0.068\\
\end{tabular}
\normalsize
\vspace{-0.1in}
\caption{Quality ablation study on Synthetic 360$^{\circ}$ scenes.}
\label{tbl:quality_ablation}
\end{table}

\begin{table}[]
\centering
\resizebox{\linewidth}{!}{
\begin{tabular}{l|cccr}
 & PSNR $\uparrow$ & SSIM $\uparrow$ & LPIPS $\downarrow$ & MB $\downarrow$ \\ \hline
SNeRG (Float) & 30.47 & 0.951 & 0.049 & 6883.6 \\
SNeRG (PNG, no $\mathcal{L}_s$) & 30.22 & 0.949 & 0.050 & 176.0 \\
SNeRG (PNG) & 30.38 & 0.950 & 0.050 & 86.7 \\
SNeRG (JPEG) & 29.71 & 0.939 & 0.062 & 70.9 \\
SNeRG (H264) & 29.86 & 0.938 & 0.065 & 30.2 \\ \hline 
JAXNeRF+ & 33.00 & 0.962 & 0.038 & 18.0 \\
JAXNeRF & 31.65 & 0.952 & 0.051 & 4.8 \\
\end{tabular}
}
\vspace{-0.1in}
\caption{Storage ablation study on Synthetic 360$^{\circ}$ scenes.}
\label{tbl:storage_ablation}
\end{table}

\begin{table}[]
\resizebox{\linewidth}{!}{
\huge
\begin{tabular}{l|cccrrc}
 & PSNR $\uparrow$ & SSIM $\uparrow$ & LPIPS $\downarrow$ & W $\downarrow$ & FPS $\uparrow$ & FPS/W $\uparrow$ \\ \hline
JAXNeRF+~\cite{jaxnerf2020github} &  {\cellcolor{red} 33.00} &  {\cellcolor{red} 0.962} &  {\cellcolor{red} 0.038} &  {300} &  {0.01} &  {0.00002} \\
NeRF~\cite{mildenhall20} &  {31.00} &  {0.947} &  {0.081} &  {300} &  {0.03} &  {0.00011} \\
JAXNeRF~\cite{jaxnerf2020github} &  {\cellcolor{yellow} 31.65} &  {\cellcolor{yellow} 0.952} &  {0.051} &  {300} &  {0.05} &  {0.00016} \\
IBRNet~\cite{wang21} &  {28.14} &  {0.942} &  {0.072} &  {300} &  {0.18} &  {0.00061} \\
AutoInt~\cite{lindell2021autoint} &  {25.55} &  {0.911} &  {0.170} &  {300} &  {0.38} &  {0.00128} \\
NSVF~\cite{liu2020nsvf} & \cellcolor{orange} 31.74 & \cellcolor{orange} 0.953 & \cellcolor{orange} 0.047 & 300 & \cellcolor{yellow} 0.65 & \cellcolor{yellow} 0.00217 \\
NV~\cite{lombardi19} &  {26.05} &  {0.893} &  {0.160} &  {300} &  {\cellcolor{orange} 3.33} &  {\cellcolor{orange} 0.01111} \\
SNeRG (PNG) & 30.38 & 0.950 & \cellcolor{yellow} 0.050 & \cellcolor{red} 85 & \cellcolor{red} 84.06 & \cellcolor{red} 0.98897 \\
\end{tabular}
}
\vspace{-0.1in}
\caption{Baseline comparisons on Synthetic 360$^{\circ}$ scenes.}
\label{tbl:synthetic_360_summary}
\end{table}

\subsection{Baseline Comparisons}

As shown in Table~\ref{tbl:synthetic_360_summary}, 
the quality of our method is comparable to all other methods, while our run-time performance is an order of magnitude faster than the fastest competing approach (Neural Volumes~\cite{lombardi19}) and more than a thousand times faster than the slowest (NeRF). 
Note that we measure the run-time rendering performance of our method on a laptop with an 85W mobile GPU, while all other methods are run on servers or workstations equipped with much more powerful GPUs (over $3\times$ the power draw).

\section{Conclusion}
We have presented a technique for rendering Neural Radiance Fields in real-time by precomputing and storing a Sparse Neural Radiance Grid. This {\em SNeRG} uses a sparse voxel grid representation to store the precomputed scene geometry, but keeps storage requirements reasonable by maintaining a neural representation for view-dependent appearance. Rendering is accelerated by evaluating the view-dependent shading network only on the visible parts of the scene, achieving over 30 frames per second on a laptop GPU for typical NeRF scenes. We hope this ability to render neural volumetric representations such as NeRF in real time on commodity graphics hardware will help increase the adoption of these neural scene representations in vision and graphics applications.

\paragraph{Acknowledgements}
We thank Keunhong Park and Michael Broxton for their generous help with debugging, and Ryan Overbeck for last-minute JavaScript help. Many thanks to John Flynn, Dominik Kaeser, Keunhong Park, Ricardo Martin-Brualla, Hugues Hoppe, Janne Kontkanen, Utkarsh Sinha, Per Karlsson, and Mark Matthews for fruitful discussions, brainstorming, and testing out our viewer.

{\small
\bibliographystyle{ieee_fullname}
\bibliography{bib}

\begin{thebibliography}{10}\itemsep=-1pt

\bibitem{Johnsson2012}
Tomas Akenine-M\"{o}ller and Bj\"{o}rn Johnsson.
\newblock Performance per what?
\newblock {\em Journal of Computer Graphics Techniques}, 2012.

\bibitem{bi2020}
Sai Bi, Zexiang Xu, Pratul~P. Srinivasan, Ben Mildenhall, Kalyan Sunkavalli,
  Miloš Hašan, Yannick Hold-Geoffroy, David Kriegman, and Ravi Ramamoorthi.
\newblock Neural reflectance fields for appearance acquisition.
\newblock {\em arXiv cs.CV arXiv:2008.03824}, 2020.

\bibitem{jax2018github}
James Bradbury, Roy Frostig, Peter Hawkins, Matthew~James Johnson, Chris Leary,
  Dougal Maclaurin, George Necula, Adam Paszke, Jake Vander{P}las, Skye
  Wanderman-{M}ilne, and Qiao Zhang.
\newblock {JAX}: composable transformations of {P}ython+{N}um{P}y programs,
  2018.
\newblock \url{http://github.com/google/jax}.

\bibitem{broxton20}
Michael Broxton, John Flynn, Ryan Overbeck, Daniel Erickson, Peter Hedman,
  Matthew DuVall, Jason Dourgarian, Jay Busch, Matt Whalen, and Paul Debevec.
\newblock Immersive light field video with a layered mesh representation.
\newblock {\em ACM Transactions on Graphics}, 2020.

\bibitem{buehler01}
Chris Buehler, Michael Bosse, Leonard McMillan, Steven Gortler, and Michael
  Cohen.
\newblock Unstructured lumigraph rendering.
\newblock {\em SIGGRAPH}, 2001.

\bibitem{chan2020pi}
Eric~R. Chan, Marco Monteiro, Petr Kellnhofer, Jiajun Wu, and Gordon Wetzstein.
\newblock pi-{GAN}: Periodic implicit generative adversarial networks for
  {3D}-aware image synthesis.
\newblock {\em CVPR}, 2021.

\bibitem{Crassin2009}
Cyril Crassin, Fabrice Neyret, Sylvain Lefebvre, and Elmar Eisemann.
\newblock {GigaVoxels}: Ray-guided streaming for efficient and detailed voxel
  rendering.
\newblock {\em Symposium on Interactive 3D Graphics and Games}, 2009.

\bibitem{davis12}
Abe Davis, Marc Levoy, and Fredo Durand.
\newblock Unstructured light fields.
\newblock {\em Computer Graphics Forum}, 2012.

\bibitem{debevec96}
Paul Debevec, C.~J. Taylor, and Jitendra Malik.
\newblock Modeling and rendering architecture from photographs: a hybrid
  geometry- and image-based approach.
\newblock {\em SIGGRAPH}, 1992.

\bibitem{Deering:1988:TPN}
Michael Deering, Stephanie Winner, Bic Schediwy, Chris Duffy, and Neil Hunt.
\newblock The triangle processor and normal vector shader: A {VLSI} system for
  high performance graphics.
\newblock {\em SIGGRAPH}, 1988.

\bibitem{jaxnerf2020github}
Boyang Deng, Jonathan~T. Barron, and Pratul~P. Srinivasan.
\newblock {JaxNeRF}: an efficient {JAX} implementation of {NeRF}, 2020.
\newblock
  \url{http://github.com/google-research/google-research/tree/master/jaxnerf}.

\bibitem{flynn19}
John Flynn, Michael Broxton, Paul Debevec, Matthew DuVall, Graham Fyffe, Ryan
  Overbeck, Noah Snavely, and Richard Tucker.
\newblock {DeepView}: View synthesis with learned gradient descent.
\newblock {\em CVPR}, 2019.

\bibitem{gafni2020dynamic}
Guy Gafni, Justus Thies, Michael Zollhöfer, and Matthias Nießner.
\newblock Dynamic neural radiance fields for monocular {4D} facial avatar
  reconstruction.
\newblock {\em CVPR}, 2021.

\bibitem{gortler96}
Steven~J. Gortler, Radek Grzeszczuk, Richard Szeliski, and Michael~F. Cohen.
\newblock The lumigraph.
\newblock {\em SIGGRAPH}, 1996.

\bibitem{hedman2018}
Peter Hedman, Julien Philip, True Price, Jan-Michael Frahm, George Drettakis,
  and Gabriel Brostow.
\newblock Deep blending for free-viewpoint image-based rendering.
\newblock {\em ACM Transactions on Graphics}, 2018.

\bibitem{adam}
Diederik~P Kingma and Jimmy Ba.
\newblock Adam: A method for stochastic optimization.
\newblock {\em ICLR}, 2015.

\bibitem{lacroute94}
Philippe Lacroute and Marc Levoy.
\newblock Fast volume rendering using a shear-warp factorization of the viewing
  transformation.
\newblock {\em SIGGRAPH}, 1994.

\bibitem{laine10}
Samuli Laine and Tero Karras.
\newblock Efficient sparse voxel octrees.
\newblock {\em I3D}, 2010.

\bibitem{lefebvre06}
Sylvain Lefebvre and Hugues Hoppe.
\newblock Perfect spatial hashing.
\newblock {\em ACM Transactions on Graphics}, 2006.

\bibitem{levoy90}
Marc Levoy.
\newblock Efficient ray tracing of volume data.
\newblock {\em ACM Transactions on Graphics}, 1980.

\bibitem{levoy96}
Marc Levoy and Pat Hanrahan.
\newblock Light field rendering.
\newblock {\em SIGGRAPH}, 1996.

\bibitem{li2021}
Zhengqi Li, Simon Niklaus, Noah Snavely, and Oliver Wang.
\newblock Neural scene flow fields for space-time view synthesis of dynamic
  scenes.
\newblock {\em CVPR}, 2021.

\bibitem{lindell2021autoint}
David~B. Lindell, Julien~N.P. Martel, and Gordon Wetzstein.
\newblock Autoint: Automatic integration for fast neural rendering.
\newblock {\em CVPR}, 2021.

\bibitem{liu2020nsvf}
Lingjie Liu, Jiatao Gu, Kyaw~Zaw Lin, Tat-Seng Chua, and Christian Theobalt.
\newblock Neural sparse voxel fields.
\newblock {\em NeurIPS}, 2020.

\bibitem{lombardi19}
Stephen Lombardi, Tomas Simon, Jason Saragih, Gabriel Schwartz, Andreas
  Lehrmann, and Yaser Sheikh.
\newblock Neural volumes: Learning dynamic renderable volumes from images.
\newblock {\em SIGGRAPH}, 2019.

\bibitem{max95}
Nelson Max.
\newblock Optical models for direct volume rendering.
\newblock {\em IEEE TVCG}, 1995.

\bibitem{waechter2014}
Michael~Goesele Michael~Waechter, Nils~Moehrle.
\newblock Let there be color! {L}arge-scale texturing of {3D} reconstructions.
\newblock {\em ECCV}, 2014.

\bibitem{mildenhall19}
Ben Mildenhall, Pratul~P. Srinivasan, Rodrigo Ortiz-Cayon, Nima~K. Kalantari,
  Ravi Ramamoorthi, Ren Ng, and Abhishek Kar.
\newblock Local light field fusion: Practical view synthesis with prescriptive
  sampling guidelines.
\newblock {\em ACM Transactions on Graphics}, 2019.

\bibitem{mildenhall20}
Ben Mildenhall, Pratul~P. Srinivasan, Matthew Tancik, Jonathan~T. Barron, Ravi
  Ramamoorthi, and Ren Ng.
\newblock {NeRF}: Representing scenes as neural radiance fields for view
  synthesis.
\newblock {\em ECCV}, 2020.

\bibitem{niessner13}
Matthias Nießner, Michael Zollhofer, Shahram Izadi, and Marc Stamminger.
\newblock Real-time {3D} reconstruction at scale using voxel hashing.
\newblock {\em ACM Transactions on Graphics}, 2013.

\bibitem{ost2020neural}
Julian Ost, Fahim Mannan, Nils Thuerey, Julian Knodt, and Felix Heide.
\newblock Neural scene graphs for dynamic scenes.
\newblock {\em CVPR}, 2021.

\bibitem{park2020deformable}
Keunhong Park, Utkarsh Sinha, Jonathan~T. Barron, Sofien Bouaziz, Dan~B
  Goldman, Steven~M. Seitz, and Ricardo Martin-Brualla.
\newblock Deformable neural radiance fields.
\newblock {\em arXiv cs.CV 2011.12948}, 2020.

\bibitem{penner17}
Eric Penner and Li Zhang.
\newblock Soft {3D} reconstruction for view synthesis.
\newblock {\em ACM Transactions on Graphics}, 2017.

\bibitem{rebain21}
Daniel Rebain, Wei Jiang, Soroosh Yazdani, Ke Li, Kwang~Moo Yi, and Andrea
  Tagliasacchi.
\newblock {DeRF}: Decomposed radiance fields.
\newblock {\em CVPR}, 2021.

\bibitem{schwarz2020graf}
Katja Schwarz, Yiyi Liao, Michael Niemeyer, and Andreas Geiger.
\newblock {GRAF}: Generative radiance fields for {3D}-aware image synthesis.
\newblock {\em NeurIPS}, 2020.

\bibitem{seitz99}
Steven~M. Seitz and Charles~R. Dyer.
\newblock Photorealistic scene reconstruction by voxel coloring.
\newblock {\em IJCV}, 1999.

\bibitem{sitzmann19}
Vincent Sitzmann, Michael Zollhoefer, and Gordon Wetzstein.
\newblock Scene representation networks: Continuous {3D}-structure-aware neural
  scene representations.
\newblock {\em NeurIPS}, 2019.

\bibitem{nerv2021}
Pratul~P. Srinivasan, Boyang Deng, Xiuming Zhang, Matthew Tancik, Ben
  Mildenhall, and Jonathan~T. Barron.
\newblock Ne{RV}: Neural reflectance and visibility fields for relighting and
  view synthesis.
\newblock {\em CVPR}, 2021.

\bibitem{srinivasan20}
Pratul~P. Srinivasan, Ben Mildenhall, Matthew Tancik, Jonathan~T. Barron,
  Richard Tucker, and Noah Snavely.
\newblock Lighthouse: Predicting lighting volumes for spatially-coherent
  illumination.
\newblock {\em CVPR}, 2020.

\bibitem{srinivasan19}
Pratul~P. Srinivasan, Richard Tucker, Jonathan~T. Barron, Ravi Ramamoorthi, Ren
  Ng, and Noah Snavely.
\newblock Pushing the boundaries of view extrapolation with multiplane images.
\newblock {\em CVPR}, 2019.

\bibitem{tancik20}
Matthew Tancik, Pratul~P. Srinivasan, Ben Mildenhall, Sara Fridovich-Keil,
  Nithin Raghavan, Utkarsh Singhal, Ravi Ramamoorthi, Jonathan~T. Barron, and
  Ren Ng.
\newblock Fourier features let networks learn high frequency functions in low
  dimensional domains.
\newblock {\em NeurIPS}, 2020.

\bibitem{thies2019neural}
Justus Thies, Michael Zollh{\"o}fer, and Matthias Nie{\ss}ner.
\newblock Deferred neural rendering: Image synthesis using neural textures.
\newblock {\em ACM Transactions on Graphics}, 2019.

\bibitem{wang21}
Qianqian Wang, Zhicheng Wang, Kyle Genova, Pratul~P. Srinivasan, Howard Zhou,
  Jonathan~T. Barron, Ricardo Martin-Brualla, Noah Snavely, and Thomas
  Funkhouser.
\newblock {IBRNet}: Learning multi-view image-based rendering.
\newblock {\em CVPR}, 2021.

\bibitem{wang2004image}
Zhou Wang, Alan~C Bovik, Hamid~R Sheikh, and Eero~P Simoncelli.
\newblock Image quality assessment: from error visibility to structural
  similarity.
\newblock {\em IEEE TIP}, 2004.

\bibitem{Westover1991}
Lee~A Westover.
\newblock Splatting: A parallel, feed-forward volume rendering algorithm.
\newblock Technical report, University of North Carolina at Chapel Hill, USA,
  1991.

\bibitem{wood00}
Daniel Wood, Daniel Azuma, Wyvern Aldinger, Brian Curless, Tom Duchamp, David
  Salesin, and Werner Stuetzle.
\newblock Surface light fields for {3D} photography.
\newblock {\em SIGGRAPH}, 2000.

\bibitem{zhang2018unreasonable}
Richard Zhang, Phillip Isola, Alexei~A Efros, Eli Shechtman, and Oliver Wang.
\newblock The unreasonable effectiveness of deep features as a perceptual
  metric.
\newblock {\em CVPR}, 2018.

\bibitem{zhou18}
Tinghui Zhou, Richard Tucker, John Flynn, Graham Fyffe, and Noah Snavely.
\newblock Stereo magnification: Learning view synthesis using multiplane
  images.
\newblock {\em ACM Transactions on Graphics}, 2018.

\end{thebibliography}
}

\appendix

\section{WebGL Implementation Details}

Our web renderer is implemented in WebGL using the THREE.js library. To conserve memory bandwidth, we load the 3D texture atlas as three separate 8-bit 3D textures: one for alpha, one for RGB and one for features. We load the indirection grid as a low resolution 8-bit 3D texture. 

During ray marching, we first query the intersection grid at the current location along the ray. If this value indicates that the macroblock is empty, we use a ray-box intersection test to skip ahead to the next macroblock along the ray.

For non-empty macroblocks, we first query the alpha texture using nearest neighbor interpolation. If alpha is zero, the current voxel within the macroblock contains empty space, and we do not fetch any additional information. If alpha is non-zero, we use trilinear interpolation to fetch the high-resolution alpha, colors and features at that voxel. This reduces the bandwidth requirement from 64 bytes per sample to 1 byte per sample for rays that are traversing empty space inside each occupied macroblock.

We implement the view-dependence MLP as simple nested for-loops in a GLSL shader. We load the network weights as 32-bit floating point textures and hard-code the network biases directly into the shader. Interestingly, we found that reducing precision lower than 32 bits did not improve the rendering performance noticeably. For added efficiency, we only evaluate the view-dependence MLP for pixels that have non-zero accumulated alpha.

\section{Performance Measurement}

We measure performance using the Chrome browser running on a 2019 MacBook Pro Laptop equipped with an 85 watt AMD Radeon Pro 5500M GPU (8GB of GPU RAM).

For accurate performance measurements, we make sure the laptop is connected to the charger, close all other applications on the laptop, and restart our browser to disable frame-rate limiting from vertical synchronization:
\begin{verbatim}
    --args --disable-gpu-vsync \
    --disable-frame-rate-limit
\end{verbatim}

In our results, we report the average frame time for rendering a 150-frame camera animation orbiting the scene (or rotating within the camera plane for forward-facing scenes). Our test-time renderings use the same image resolutions and camera intrinsics as the input training images:
\begin{itemize}
    \item $39^\circ$ field-of-view at $800 \times 800$ for Synthetic 360$^\circ$ scenes,
    \item $53^\circ$ field-of-view at $1006 \times 756$ for Real Forward-Facing, and
    \item $53^\circ$ at field-of-view $990 \times 773$ for Real 360$^\circ$ scenes.
\end{itemize}

\section{Additional Experiment Details}

\subsection{Experiments with Changing 3D Resolution}

Table~\ref{tbl:res} demonstrates that our method is able to achieve even higher rendering speeds and lower storage costs by baking the 3D grids at a lower resolution, at the expense of a slight decrease in rendering quality.

\begin{table}[]
\centering
\resizebox{\linewidth}{!}{
\begin{tabular}{l|ccccrr}
 & PSNR $\uparrow$ & SSIM $\uparrow$ & LPIPS $\downarrow$ & FPS $\uparrow$ & MB $\downarrow$ \\ \hline 
 $1000^3$ & 30.38 & 0.950 & 0.050 & 84.06 & 86.7\\
 $750^3$ & 29.94 & 0.947 & 0.054 & 89.82 & 44.7\\
 $500^3$ & 28.93 & 0.939 & 0.064 & 101.78 & 17.8\\
\end{tabular}
}
\caption{Voxel grid resolution ablation using SNeRG (PNG) on Synthetic $360^\circ$ Scenes.}
\label{tbl:res}
\end{table}

\subsection{Real 360$^\circ$ Scenes}

We evaluate our method on the two real 360$^\circ$ scenes provided by the original NeRF paper (\scenename{Flowers} and \scenename{Pine Cone}) and two new scenes that we have captured ourselves (\scenename{Toy Car} and \scenename{Spheres}). All four datasets contain 100-200 images where the camera orbits around an object. Note that the \scenename{Spheres} scene contains glossy objects that are hard to model using diffuse geometry alone.

Tables~\ref{tbl:real360_psnr},~\ref{tbl:real360_ssim}, and~\ref{tbl:real360_lpips} demonstrate that our method is able to maintain rendering quality close to the trained NeRF models while rendering about 30 frames per second (Table~\ref{tbl:real360_performance}). Table~\ref{tbl:storage} studies the impact of using different image and video compression algorithms for these datasets, and shows that we are able to store these scenes using about 50 MB.

We train all NeRF models on this data by shifting and scaling the camera translations so that they approximately lie on a sphere around the origin, and sampling points linearly in disparity along each camera ray, as done by Mildenhall \etal. After training, we manually set a bounding box to isolate the objects of interest in the scene and ignore the unbounded peripheral content that is not sampled well enough for NeRF to recover. During baking, we only evaluate the subset of the scene which is inside this bounding box. We change our quality measurements to reflect this, masking all of the images (our results, baseline results, and ground truth images) using the alpha mask generated by our method. Otherwise, the results would be significantly biased by the missing background geometry that was outside the scene bounding box. Interestingly, we find that a diffuse-only model without any view-dependent effects is surprisingly competitive for these scenes, potentially due to the low-frequency lighting conditions during capture. Additionally, the diffuse model is able to reasonably fake view-dependent effects in some cases by hiding mirrored versions of reflected content inside the objects' surfaces.

\begin{table}[]
\centering
\resizebox{\linewidth}{!}{
\huge
\begin{tabular}{l|cccccc}
&  &  & \scenename{Toy} & & \scenename{Pine } \\ 
& Mean & \scenename{Spheres} & \scenename{Car} & \scenename{Flowers} & \scenename{Cone} \\ \hline 
JAXNeRF+& \multicolumn{1}{c}{\cellcolor{red} 24.56} & \multicolumn{1}{c}{\cellcolor{red} 23.56} & \multicolumn{1}{c}{26.13} & \multicolumn{1}{c}{\cellcolor{red} 25.22} & \multicolumn{1}{c}{\cellcolor{red} 23.33}\\
JAXNeRF+ Deferred& \multicolumn{1}{c}{\cellcolor{orange} 24.31} & \multicolumn{1}{c}{\cellcolor{orange} 23.44} & \multicolumn{1}{c}{\cellcolor{yellow} 26.16} & \multicolumn{1}{c}{\cellcolor{orange} 24.81} & \multicolumn{1}{c}{\cellcolor{orange} 22.81}\\
SNeRG (PNG)& \multicolumn{1}{c}{23.97} & \multicolumn{1}{c}{23.16} & \multicolumn{1}{c}{\cellcolor{orange} 26.17} & \multicolumn{1}{c}{\cellcolor{yellow} 24.43} & \multicolumn{1}{c}{22.14}\\
JAXNeRF+ Diffuse& \multicolumn{1}{c}{\cellcolor{yellow} 24.02} & \multicolumn{1}{c}{\cellcolor{yellow} 23.26} & \multicolumn{1}{c}{\cellcolor{red} 26.17} & \multicolumn{1}{c}{24.23} & \multicolumn{1}{c}{\cellcolor{yellow} 22.42}\\
\end{tabular}
} \caption{PSNR $\uparrow$, Real 360$^{\circ}$ scenes.}
\label{tbl:real360_psnr}
\end{table}

\begin{table}[]
\centering
\resizebox{\linewidth}{!}{
\huge
\begin{tabular}{l|cccccc}
&  &  & \scenename{Toy} & & \scenename{Pine } \\ 
& Mean & \scenename{Spheres} & \scenename{Car} & \scenename{Flowers} & \scenename{Cone} \\ \hline 
JAXNeRF+& \multicolumn{1}{c}{\cellcolor{red} 0.703} & \multicolumn{1}{c}{\cellcolor{red} 0.573} & \multicolumn{1}{c}{\cellcolor{red} 0.792} & \multicolumn{1}{c}{\cellcolor{red} 0.765} & \multicolumn{1}{c}{\cellcolor{red} 0.681}\\
JAXNeRF+ Deferred& \multicolumn{1}{c}{\cellcolor{orange} 0.693} & \multicolumn{1}{c}{\cellcolor{yellow} 0.563} & \multicolumn{1}{c}{0.787} & \multicolumn{1}{c}{\cellcolor{orange} 0.754} & \multicolumn{1}{c}{\cellcolor{orange} 0.666}\\
SNeRG (PNG)& \multicolumn{1}{c}{0.662} & \multicolumn{1}{c}{0.521} & \multicolumn{1}{c}{\cellcolor{orange} 0.788} & \multicolumn{1}{c}{\cellcolor{yellow} 0.746} & \multicolumn{1}{c}{0.595}\\
JAXNeRF+ Diffuse& \multicolumn{1}{c}{\cellcolor{yellow} 0.681} & \multicolumn{1}{c}{\cellcolor{orange} 0.565} & \multicolumn{1}{c}{\cellcolor{yellow} 0.788} & \multicolumn{1}{c}{0.728} & \multicolumn{1}{c}{\cellcolor{yellow} 0.645}\\
\end{tabular}
} \caption{SSIM $\uparrow$, Real 360$^{\circ}$ scenes.}
\label{tbl:real360_ssim}
\end{table}

\begin{table}[]
\centering
\resizebox{\linewidth}{!}{
\huge
\begin{tabular}{l|cccccc}
&  &  & \scenename{Toy} & & \scenename{Pine } \\ 
& Mean & \scenename{Spheres} & \scenename{Car} & \scenename{Flowers} & \scenename{Cone} \\ \hline 
JAXNeRF+& \multicolumn{1}{c}{\cellcolor{red} 0.248} & \multicolumn{1}{c}{\cellcolor{orange} 0.311} & \multicolumn{1}{c}{\cellcolor{red} 0.195} & \multicolumn{1}{c}{\cellcolor{red} 0.227} & \multicolumn{1}{c}{\cellcolor{red} 0.257}\\
JAXNeRF+ Deferred& \multicolumn{1}{c}{\cellcolor{yellow} 0.261} & \multicolumn{1}{c}{\cellcolor{yellow} 0.320} & \multicolumn{1}{c}{\cellcolor{yellow} 0.209} & \multicolumn{1}{c}{\cellcolor{orange} 0.244} & \multicolumn{1}{c}{\cellcolor{orange} 0.271}\\
SNeRG (PNG)& \multicolumn{1}{c}{0.293} & \multicolumn{1}{c}{0.357} & \multicolumn{1}{c}{0.209} & \multicolumn{1}{c}{0.272} & \multicolumn{1}{c}{0.336}\\
JAXNeRF+ Diffuse& \multicolumn{1}{c}{\cellcolor{orange} 0.260} & \multicolumn{1}{c}{\cellcolor{red} 0.306} & \multicolumn{1}{c}{\cellcolor{orange} 0.200} & \multicolumn{1}{c}{\cellcolor{yellow} 0.257} & \multicolumn{1}{c}{\cellcolor{yellow} 0.277}\\
\end{tabular}
} \caption{LPIPS $\downarrow$, Real 360$^{\circ}$ scenes.}
\label{tbl:real360_lpips}
\end{table}

\begin{table}[]
\centering
\begin{tabular}{cccc} & \scenename{Toy} & & \scenename{Pine } \\ 
 \scenename{Spheres} & \scenename{Car} & \scenename{Flowers} & \scenename{Cone} \\ \hline 
53.7 & 41.1 & 40.8 & 39.5 \\
\end{tabular}
\caption{Performance (FPS $\uparrow$), Real $360^\circ$ Scenes.}
\label{tbl:real360_performance}
\end{table}

\begin{figure}
\centering
\begin{tabular}{@{}c@{\,}c@{}}
    \includegraphics[width=0.49\columnwidth]{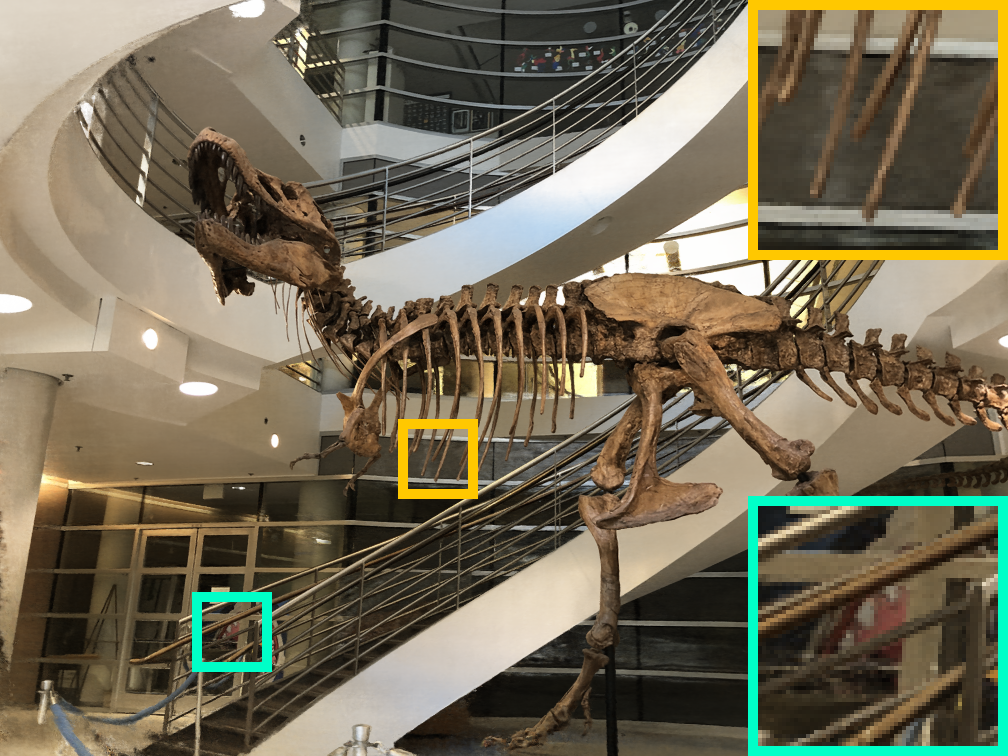} & 
    \includegraphics[width=0.49\columnwidth]{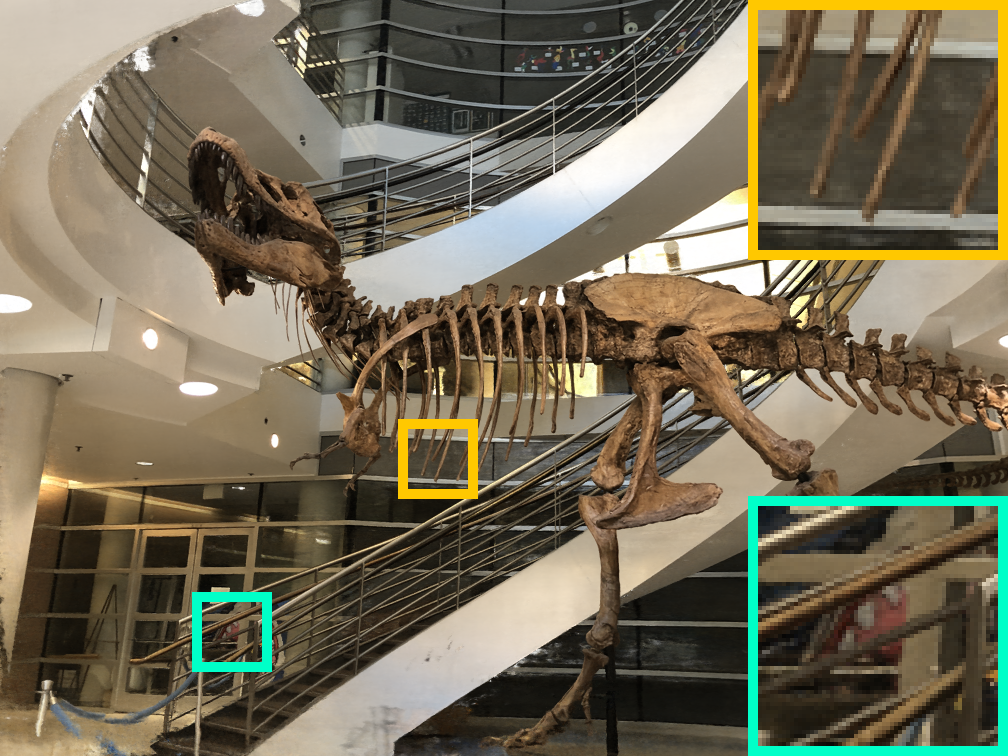} \\
    \small (a) JAXNeRF+ (25.50) & 
    \small (b) Deferred (24.91) \\ 
    \includegraphics[width=0.49\columnwidth]{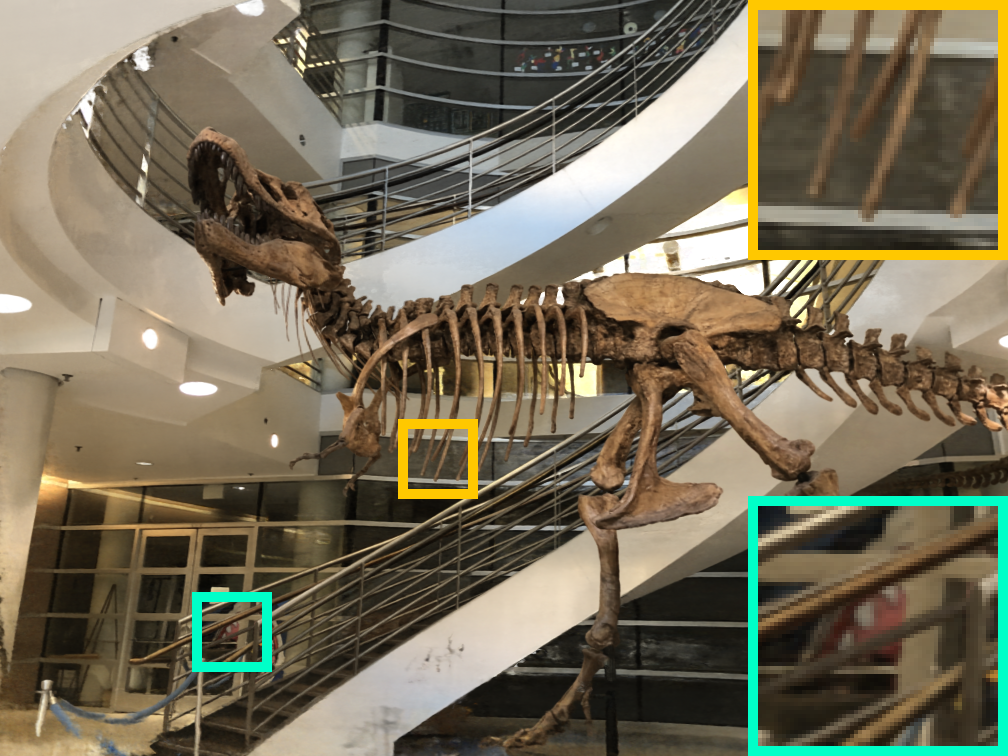} & 
    \includegraphics[width=0.49\columnwidth]{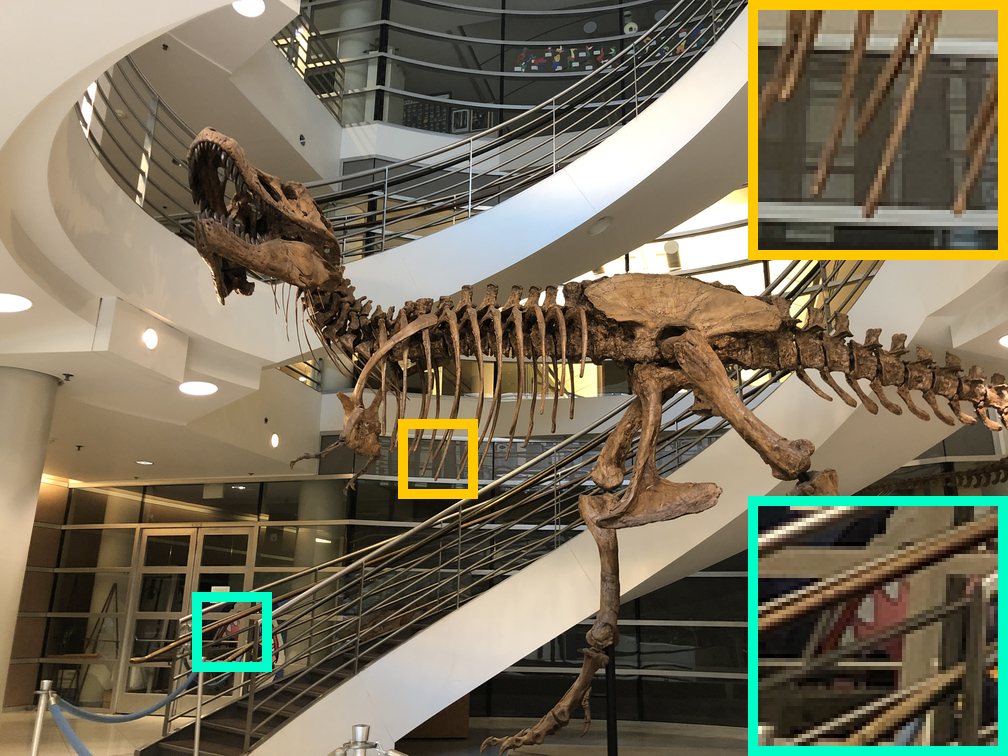} \\
    \small (c) SNeRG (24.43) & 
    \small (d) Ground Truth \\ 
    \end{tabular}
        \caption{\textbf{Real forward-facing scene example results} (PSNR in parentheses).}
        \label{fig:realff}
\end{figure}

\subsection{Real Forward-Facing Scenes}

\begin{table}[]
\resizebox{\linewidth}{!}{
\begin{tabular}{l|cccrrc}
 & PSNR $\uparrow$ & SSIM $\uparrow$ & LPIPS $\downarrow$ & W $\downarrow$ & FPS $\uparrow$ & FPS/W $\uparrow$ \\ \hline 
JAXNeRF+ & \cellcolor{red} 26.95 & \cellcolor{orange} 0.845 & \cellcolor{red} 0.145 & 300 & 0.00 & 0.00001 \\
DeRF & 24.81 & 0.767 & 0.274 & 300 & 0.03 & 0.00009 \\
NeRF & 26.50 & 0.811 & 0.250 & 300 & 0.03 & 0.00011 \\
JAXNeRF & \cellcolor{orange} 26.92 & \cellcolor{yellow} 0.831 & \cellcolor{orange} 0.173 & 300 & 0.04 & 0.00013 \\
IBRNet & \cellcolor{yellow} 26.73 & \cellcolor{red} 0.851 & \cellcolor{yellow} 0.175 & 300 & \cellcolor{yellow} 0.18 & \cellcolor{yellow} 0.00061 \\
LLFF & 24.13 & 0.798 & 0.212 & 250 & \cellcolor{red} 60.00 & \cellcolor{red} 0.24000 \\
SNeRG (PNG) & 25.63 & 0.818 & 0.183 & \cellcolor{red} 85 & \cellcolor{orange} 27.38 & \cellcolor{red} 0.32210 \\
\end{tabular}
}
\vspace{-0.1in}
\caption{Quality and performance comparison for Real Forward-Facing scenes.}
\label{tbl:forward_facing_summary}
\end{table}

\begin{table}[]
\resizebox{\linewidth}{!}{
\centering
\begin{tabular}{cccccccc}
\scenename{T-Rex} & \scenename{Leaves} & \scenename{Room} & \scenename{Orchids} & \scenename{Horns} & \scenename{Fortress} & \scenename{Fern} & \scenename{Flower} \\ \hline 
34.78 & 19.02 & 37.09 & 18.54 & 34.02 & 39.16 & 26.73 & 26.58 \\
\end{tabular}
}
\caption{Performance (FPS $\uparrow$), Real Forward-Facing Scenes.}
\label{tbl:forward_facing_fps}
\vspace{-0.1in}
\end{table}

We also evaluate our approach on the real forward-facing scenes in the NeRF paper (Tables~\ref{tbl:forward_facing_summary},~\ref{tbl:forward_facing_fps}, and~\ref{tbl:storage}). Since these scenes are only captured and viewed from a limited range of forward-facing viewpoints, layered representations such as multi-plane images~\cite{flynn19, mildenhall19, penner17, srinivasan19, zhou18} are a compelling option for real-time rendering. Note that the normalized device coordinate transformation used in NeRF for these forward-facing scenes can be interpreted as transforming NeRF into a continuous version of a multiplane image representation that supports larger viewpoint changes.

We found that our baking procedure sometimes reduces the total alpha mass in the scene, introducing small semi-transparent holes for these datasets. To overcome this, we partially un-premultiply alpha after ray marching. That is, after alpha compositing:
\begin{equation}
    \text{rgbafeatures} \leftarrow \text{rgbafeatures} \times \frac{\min(1.0, 1.5\alpha)}{\alpha}
\end{equation}
if $\alpha > 0$. This fully saturates alpha values above 0.66, while still allowing for soft edges and a smooth fall-off.

\subsection{Baselines}

Here we provide additional details for the baseline methods we use in our experiments.

\shortpara{NeRF~\cite{mildenhall20}}
We directly use the results reported in the original paper by Mildenhall \etal. Run-time was measured on a single NVIDIA V100 GPU.

\shortpara{JAXNeRF~\cite{jaxnerf2020github}} is a JAX implementation of NeRF, with default settings (64 + 128 samples per ray, MLP width of 256). Run-time was measured on an NVIDIA V100 GPU.

\shortpara{JAXNeRF+} is a more compute-intensive version of JAXNeRF, trained with 192 + 384 samples per ray and an MLP width of 512 channels. Run-time was measured on a single NVIDIA V100 GPU. We use this architecture as a starting point for our modifications (deferred shading and baking), as using more samples per ray allows us to recover a sparser representation that better concentrates opacity near object surfaces.

\shortpara{JAXNeRF+ Tinyview}
This baseline measures the effects of using a smaller network architecture (same as $\mlp_\modelweightsspec$) for the view-dependent effects. It uses the same architecture for the view-dependent effects as our ``Deferred'' model, but evaluates view-dependent effects for every 3D sample instead of once per pixel.

\shortpara{JAXNeRF+ Diffuse}
This baseline measures the effects of modeling view-dependent appearance. It uses the same architecture as JAXNeRF+, but replaces the view dependence network with a single layer that directly outputs a color without any knowledge of the viewing direction.

\shortpara{AutoInt~\cite{lindell2021autoint}}
We use the N=8 setting reported by Lindell \etal, which achieves their highest ratio of quality to run-time. The authors did not mention what hardware they ran on, but we are assuming that they also run on an NVIDIA V100 GPU since they directly compare to NeRF runtimes.

\shortpara{Neural Volumes~\cite{lombardi19}}
We copy the rendering quality results reported in the NeRF paper and copy the rendering run-time results reported in the AutoInt paper. We assume that the run-times reported in the AutoInt paper are measured on an NVIDIA V100 GPU since the AutoInt paper directly compares these results with NeRF run-times.

\shortpara{NSVF~\cite{liu2020nsvf}}
We use the average run-time of 1.537 seconds per frame reported by the authors, using early stopping. Performance was measured on an NVIDIA V100 GPU.

\shortpara{DeRF~\cite{rebain21}}
We use the DeRF model with 8 heads and 96 channels per head, which achieves the highest ratio of quality to run-time according to the results in their paper. Run-times were measured on an NVIDIA V100 GPU.

\shortpara{IBRNet~\cite{wang21}}
We use the highest quality results (per-scene fine-tuned) in their paper. Run-times were estimated by scaling the NVIDIA V100 GPU NeRF run-times according to the TFLOPs in Table 3 of their paper.

\shortpara{LLFF~\cite{mildenhall19}}
Run-times were measured using the original CUDA implementation on a GTX 2080 Ti (250W).

\subsection{Experiments with Changing 3D Resolution}

Table~\ref{tbl:res} demonstrates that our method is able to achieve even higher rendering speeds and lower storage costs by baking the 3D grids at a lower resolution, at the expense of a slight decrease in rendering quality.

\subsection{Per-Scene Quality and Performance Metrics}

Tables~\ref{tbl:per_scene_synth_psnr}-\ref{tbl:per_scene_synth_lpips}, provide a per-scene breakdown for the quality metrics in the Synthetic $360^\circ$ scenes. Similar breakdowns for the Real Forward Facing scene can be found in Tables~\ref{tbl:per_scene_real_psnr}-\ref{tbl:per_scene_real_lpips}. Table~\ref{tbl:per_scene_time} shows the per-scene frame time and Table~\ref{tbl:per_scene_mem} shows the per-scene GPU memory consumption our performance ablations: 1) removing the view-dependence MLP, 2) removing the sparsity loss, and 3) switching from `deferred'' rendering back to querying an MLP at each sample along the ray.

\begin{table*}[]
\centering
\resizebox{\linewidth}{!}{
\begin{tabular}{@{}l|ccc@{\,\,\,}r|ccc@{\,\,\,}r|ccc@{\,\,\,}r@{}}
 & \multicolumn{4}{c|}{Synthetic 360$^\circ$}
 & \multicolumn{4}{c|}{Real Forward-Facing}
 & \multicolumn{4}{c}{Real 360$^\circ$}\\
 & PSNR $\uparrow$ & SSIM $\uparrow$ & LPIPS $\downarrow$ & MB $\downarrow$ & PSNR $\uparrow$ & SSIM $\uparrow$ & LPIPS $\downarrow$ & MB $\downarrow$ & PSNR $\uparrow$ & SSIM $\uparrow$ & LPIPS $\downarrow$ & MB $\downarrow$\\ \hline
SNeRG (Float) & 30.47 & 0.951 & 0.049 & 6919.9  & 25.74 & 0.823 & 0.180 & 13830.3  & 24.05 & 0.661 & 0.299 & 7238.3 \\
SNeRG (PNG) & 30.38 & 0.950 & 0.050 & 86.7  & 25.63 & 0.818 & 0.183 & 373.2  & 23.97 & 0.662 & 0.293 & 264.7 \\
SNeRG (JPG) & 29.71 & 0.939 & 0.062 & 70.9  & 25.27 & 0.781 & 0.232 & 183.3  & 23.67 & 0.638 & 0.306 & 129.2 \\
SNeRG (H264) & 29.86 & 0.938 & 0.065 & 30.2  & 25.13 & 0.761 & 0.257 & 66.9  & 23.60 & 0.629 & 0.316 & 51.3 \\ \hline 
JAXNeRF+ & 33.00 & 0.962 & 0.038 & 18.0  & 26.95 & 0.845 & 0.145 & 18.0  & 24.56 & 0.703 & 0.248 & 18.0 \\
\end{tabular}
}
\caption{Storage ablation.}
\label{tbl:storage}
\end{table*}

\begin{table*}[]
\centering
\begin{tabular}{l|ccccccccc}
& \multicolumn{9}{c}{PSNR $\uparrow$} \\
& Mean & \scenename{Chair} & \scenename{Drums} & \scenename{Ficus} & \scenename{Hotdog} & \scenename{Lego} & \scenename{Materials} & \scenename{Mic} & \scenename{Ship} \\ \hline 
AutoInt&  {25.55} &  {25.60} &  {20.78} &  {22.47} &  {32.33} &  {25.09} &  {25.90} &  {28.10} &  {24.15}\\
NV&  {26.05} &  {28.33} &  {22.58} &  {24.79} &  {30.71} &  {26.08} &  {24.22} &  {27.78} &  {23.93}\\
IBRNet&  {28.14} & --- & --- & --- & --- & --- & --- & --- & ---\\
NeRF&  {31.00} &  {33.00} &  {25.01} &  {30.13} &  {36.18} &  {32.54} &  {29.62} &  {32.91} &  {28.65}\\
JAXNeRF&  {\cellcolor{yellow} 31.65} &  {\cellcolor{yellow} 33.88} &  {\cellcolor{yellow} 25.08} &  {\cellcolor{yellow} 30.51} &  {\cellcolor{yellow} 36.91} &  {33.24} &  {\cellcolor{yellow} 30.03} &  {\cellcolor{orange} 34.52} &  {\cellcolor{yellow} 29.07}\\
NSVF&  {\cellcolor{orange} 31.74} &  {33.19} &  {\cellcolor{orange} 25.18} &  {\cellcolor{orange} 31.23} &  {\cellcolor{orange} 37.14} &  {32.29} &  {\cellcolor{red} 32.68} &  {34.27} &  {27.93}\\
JAXNeRF+&  {\cellcolor{red} 33.00} &  {\cellcolor{red} 35.35} &  {\cellcolor{red} 25.65} &  {\cellcolor{red} 32.77} &  {\cellcolor{red} 37.58} &  {\cellcolor{red} 35.35} &  {\cellcolor{orange} 30.29} &  {\cellcolor{red} 36.52} &  {\cellcolor{red} 30.48}\\ \hline 
JAXNeRF+ Tinyview&  {31.65} &  {\cellcolor{orange} 34.24} &  {25.06} &  {29.52} &  {36.75} &  {\cellcolor{yellow} 34.34} &  {29.17} &  {\cellcolor{yellow} 34.31} &  {\cellcolor{orange} 29.84}\\
JAXNeRF+ Deferred&  {30.55} &  {33.63} &  {23.73} &  {28.46} &  {35.10} &  {\cellcolor{orange} 34.67} &  {26.74} &  {33.03} &  {29.04}\\
SNeRG (PNG)&  {30.38} &  {33.24} &  {24.57} &  {29.32} &  {34.33} &  {33.82} &  {27.21} &  {32.60} &  {27.97}\\
JAXNeRF+ Diffuse&  {27.39} &  {29.95} &  {21.93} &  {22.37} &  {32.99} &  {32.17} &  {24.83} &  {28.36} &  {26.57}\\
\end{tabular}
\caption{PSNR, Synthetic 360$^{\circ}$ scenes.}
\label{tbl:per_scene_synth_psnr}
\end{table*}

\begin{table*}[]
\centering
\begin{tabular}{l|ccccccccc}
& \multicolumn{9}{c}{SSIM $\uparrow$} \\
& Mean & \scenename{Chair} & \scenename{Drums} & \scenename{Ficus} & \scenename{Hotdog} & \scenename{Lego} & \scenename{Materials} & \scenename{Mic} & \scenename{Ship} \\ \hline 
AutoInt&  {0.911} &  {0.928} &  {0.861} &  {0.898} &  {0.974} &  {0.900} &  {0.930} &  {0.948} &  {0.852}\\
NV&  {0.893} &  {0.916} &  {0.873} &  {0.910} &  {0.944} &  {0.880} &  {0.888} &  {0.946} &  {0.784}\\
IBRNet&  {0.942} & --- & --- & --- & --- & --- & --- & --- & ---\\
NeRF&  {0.947} &  {0.967} &  {0.925} &  {0.964} &  {0.974} &  {0.961} &  {0.949} &  {0.980} &  {0.856}\\
JAXNeRF&  {0.952} &  {0.974} &  {0.927} &  {0.967} &  {0.979} &  {0.968} &  {\cellcolor{yellow} 0.952} &  {\cellcolor{yellow} 0.987} &  {0.865}\\
NSVF&  {\cellcolor{yellow} 0.953} &  {0.968} &  {\cellcolor{orange} 0.931} &  {\cellcolor{orange} 0.973} &  {\cellcolor{orange} 0.980} &  {0.960} &  {\cellcolor{red} 0.973} &  {\cellcolor{orange} 0.987} &  {0.854}\\
JAXNeRF+&  {\cellcolor{red} 0.962} &  {\cellcolor{red} 0.982} &  {\cellcolor{red} 0.936} &  {\cellcolor{red} 0.980} &  {\cellcolor{red} 0.983} &  {\cellcolor{red} 0.979} &  {\cellcolor{orange} 0.956} &  {\cellcolor{red} 0.991} &  {\cellcolor{red} 0.887}\\ \hline 
JAXNeRF+ Tinyview&  {\cellcolor{orange} 0.954} &  {\cellcolor{orange} 0.978} &  {0.925} &  {0.966} &  {\cellcolor{yellow} 0.979} &  {\cellcolor{yellow} 0.975} &  {0.946} &  {0.986} &  {\cellcolor{orange} 0.880}\\
JAXNeRF+ Deferred&  {0.952} &  {\cellcolor{yellow} 0.976} &  {0.922} &  {0.964} &  {0.976} &  {\cellcolor{orange} 0.976} &  {0.939} &  {0.984} &  {\cellcolor{yellow} 0.874}\\
SNeRG (PNG)&  {0.950} &  {0.975} &  {\cellcolor{yellow} 0.929} &  {\cellcolor{yellow} 0.967} &  {0.971} &  {0.973} &  {0.938} &  {0.982} &  {0.865}\\
JAXNeRF+ Diffuse&  {0.927} &  {0.951} &  {0.888} &  {0.916} &  {0.966} &  {0.968} &  {0.911} &  {0.967} &  {0.850}\\
\end{tabular}
\caption{SSIM, Synthetic 360$^{\circ}$ scenes.}
\label{tbl:per_scene_synth_ssim}
\end{table*}

\begin{table*}[]
\centering
\begin{tabular}{l|ccccccccc}
& \multicolumn{9}{c}{LPIPS $\downarrow$} \\
        & Mean & \scenename{Chair} & \scenename{Drums} & \scenename{Ficus} & \scenename{Hotdog} & \scenename{Lego} & \scenename{Materials} & \scenename{Mic} & \scenename{Ship} \\ \hline 
AutoInt&  {0.170} &  {0.141} &  {0.224} &  {0.148} &  {0.080} &  {0.175} &  {0.136} &  {0.131} &  {0.323}\\
NV&  {0.160} &  {0.109} &  {0.214} &  {0.162} &  {0.109} &  {0.175} &  {0.130} &  {0.107} &  {0.276}\\
IBRNet&  {0.072} & --- & --- & --- & --- & --- & --- & --- & ---\\
NeRF&  {0.081} &  {0.046} &  {0.091} &  {0.044} &  {0.121} &  {0.050} &  {0.063} &  {0.028} &  {0.206}\\
JAXNeRF&  {0.051} &  {0.027} &  {0.070} &  {0.033} &  {0.030} &  {0.030} &  {\cellcolor{yellow} 0.048} &  {\cellcolor{yellow} 0.013} &  {0.156}\\
NSVF&  {\cellcolor{yellow} 0.047} &  {0.043} &  {0.069} &  {\cellcolor{red} 0.017} &  {\cellcolor{orange} 0.025} &  {0.029} &  {\cellcolor{red} 0.021} &  {\cellcolor{orange} 0.010} &  {0.162}\\
JAXNeRF+&  {\cellcolor{red} 0.038} &  {\cellcolor{red} 0.017} &  {\cellcolor{red} 0.057} &  {\cellcolor{orange} 0.018} &  {\cellcolor{red} 0.022} &  {\cellcolor{red} 0.017} &  {\cellcolor{orange} 0.041} &  {\cellcolor{red} 0.008} &  {\cellcolor{red} 0.123}\\ \hline 
JAXNeRF+ Tinyview&  {\cellcolor{orange} 0.047} &  {\cellcolor{orange} 0.020} &  {0.079} &  {0.030} &  {\cellcolor{yellow} 0.028} &  {\cellcolor{yellow} 0.020} &  {0.051} &  {0.016} &  {\cellcolor{orange} 0.130}\\
JAXNeRF+ Deferred&  {0.049} &  {\cellcolor{yellow} 0.022} &  {\cellcolor{yellow} 0.069} &  {0.041} &  {0.033} &  {\cellcolor{orange} 0.019} &  {0.052} &  {0.016} &  {\cellcolor{yellow} 0.138}\\
SNeRG (PNG)&  {0.050} &  {0.025} &  {\cellcolor{orange} 0.061} &  {\cellcolor{yellow} 0.028} &  {0.043} &  {0.022} &  {0.052} &  {0.016} &  {0.156}\\
JAXNeRF+ Diffuse&  {0.068} &  {0.048} &  {0.101} &  {0.074} &  {0.044} &  {0.024} &  {0.074} &  {0.031} &  {0.152}\\
\end{tabular}
\caption{LPIPS, Synthetic 360$^{\circ}$ scenes.}
\label{tbl:per_scene_synth_lpips}
\end{table*}

\begin{table*}[]
\centering
\begin{tabular}{l|ccccccccc}
& \multicolumn{9}{c}{PSNR $\uparrow$} \\
        & Mean & \scenename{Room} & \scenename{Fern} & \scenename{Leaves} & \scenename{Fortress} & \scenename{Orchids} & \scenename{Flower} & \scenename{T-Rex} & \scenename{Horns} \\ \hline 
LLFF& 24.13 & \multicolumn{1}{c}{28.42} & \multicolumn{1}{c}{22.85} & \multicolumn{1}{c}{19.52} & \multicolumn{1}{c}{29.40} & \multicolumn{1}{c}{18.52} & \multicolumn{1}{c}{25.46} & \multicolumn{1}{c}{24.15} & \multicolumn{1}{c}{24.70}\\
DeRF& 24.81 & \multicolumn{1}{c}{29.72} & \multicolumn{1}{c}{24.87} & \multicolumn{1}{c}{20.64} & \multicolumn{1}{c}{26.84} & \multicolumn{1}{c}{19.97} & \multicolumn{1}{c}{25.66} & \multicolumn{1}{c}{24.86} & \multicolumn{1}{c}{25.89}\\
NeRF& 26.50 & \multicolumn{1}{c}{\cellcolor{yellow} 32.70} & \multicolumn{1}{c}{\cellcolor{red} 25.17} & \multicolumn{1}{c}{\cellcolor{orange} 20.92} & \multicolumn{1}{c}{\cellcolor{yellow} 31.16} & \multicolumn{1}{c}{\cellcolor{red} 20.36} & \multicolumn{1}{c}{\cellcolor{yellow} 27.40} & \multicolumn{1}{c}{26.80} & \multicolumn{1}{c}{27.45}\\
IBRNet& \cellcolor{yellow} 26.73 & --- & --- & --- & --- & --- & --- & --- & ---\\
JAXNeRF& \cellcolor{orange} 26.92 & \multicolumn{1}{c}{\cellcolor{orange} 33.30} & \multicolumn{1}{c}{\cellcolor{yellow} 24.92} & \multicolumn{1}{c}{\cellcolor{red} 21.24} & \multicolumn{1}{c}{\cellcolor{red} 31.78} & \multicolumn{1}{c}{\cellcolor{orange} 20.32} & \multicolumn{1}{c}{\cellcolor{orange} 28.09} & \multicolumn{1}{c}{27.43} & \multicolumn{1}{c}{\cellcolor{yellow} 28.29}\\
JAXNeRF+& \cellcolor{red} 26.95 & \multicolumn{1}{c}{\cellcolor{red} 33.79} & \multicolumn{1}{c}{24.38} & \multicolumn{1}{c}{\cellcolor{yellow} 20.82} & \multicolumn{1}{c}{31.14} & \multicolumn{1}{c}{\cellcolor{yellow} 20.09} & \multicolumn{1}{c}{\cellcolor{red} 28.34} & \multicolumn{1}{c}{\cellcolor{orange} 27.94} & \multicolumn{1}{c}{\cellcolor{red} 29.08}\\ \hline 
JAXNeRF+ Deferred& 26.61 & \multicolumn{1}{c}{32.63} & \multicolumn{1}{c}{24.88} & \multicolumn{1}{c}{20.67} & \multicolumn{1}{c}{\cellcolor{orange} 31.28} & \multicolumn{1}{c}{19.72} & \multicolumn{1}{c}{27.40} & \multicolumn{1}{c}{\cellcolor{yellow} 27.72} & \multicolumn{1}{c}{\cellcolor{orange} 28.56}\\
SNeRG (PNG)& 25.63 & \multicolumn{1}{c}{30.04} & \multicolumn{1}{c}{24.85} & \multicolumn{1}{c}{20.01} & \multicolumn{1}{c}{30.91} & \multicolumn{1}{c}{19.73} & \multicolumn{1}{c}{27.00} & \multicolumn{1}{c}{25.80} & \multicolumn{1}{c}{26.71}\\
JAXNeRF+ Diffuse& 26.31 & \multicolumn{1}{c}{31.44} & \multicolumn{1}{c}{\cellcolor{orange} 24.98} & \multicolumn{1}{c}{20.64} & \multicolumn{1}{c}{30.46} & \multicolumn{1}{c}{19.89} & \multicolumn{1}{c}{26.95} & \multicolumn{1}{c}{\cellcolor{red} 28.06} & \multicolumn{1}{c}{28.03}\\
\end{tabular}
\caption{PSNR, Real Forward-Facing scenes.}
\label{tbl:per_scene_real_psnr}
\end{table*}

\begin{table*}[]
\centering
\begin{tabular}{l|ccccccccc}
& \multicolumn{9}{c}{SSIM $\uparrow$} \\
        & Mean & \scenename{Room} & \scenename{Fern} & \scenename{Leaves} & \scenename{Fortress} & \scenename{Orchids} & \scenename{Flower} & \scenename{T-Rex} & \scenename{Horns} \\ \hline 
LLFF& 0.798 & \multicolumn{1}{c}{0.932} & \multicolumn{1}{c}{0.753} & \multicolumn{1}{c}{0.697} & \multicolumn{1}{c}{0.872} & \multicolumn{1}{c}{0.588} & \multicolumn{1}{c}{0.844} & \multicolumn{1}{c}{0.857} & \multicolumn{1}{c}{0.840}\\
DeRF& 0.767 & \multicolumn{1}{c}{0.930} & \multicolumn{1}{c}{0.770} & \multicolumn{1}{c}{0.680} & \multicolumn{1}{c}{0.730} & \multicolumn{1}{c}{0.610} & \multicolumn{1}{c}{0.790} & \multicolumn{1}{c}{0.840} & \multicolumn{1}{c}{0.790}\\
NeRF& 0.811 & \multicolumn{1}{c}{0.948} & \multicolumn{1}{c}{0.792} & \multicolumn{1}{c}{0.690} & \multicolumn{1}{c}{0.881} & \multicolumn{1}{c}{0.641} & \multicolumn{1}{c}{0.827} & \multicolumn{1}{c}{0.880} & \multicolumn{1}{c}{0.828}\\
IBRNet& \cellcolor{red} 0.851 & --- & --- & --- & --- & --- & --- & --- & ---\\
JAXNeRF& 0.831 & \multicolumn{1}{c}{\cellcolor{orange} 0.958} & \multicolumn{1}{c}{0.806} & \multicolumn{1}{c}{\cellcolor{yellow} 0.717} & \multicolumn{1}{c}{\cellcolor{yellow} 0.897} & \multicolumn{1}{c}{\cellcolor{orange} 0.657} & \multicolumn{1}{c}{\cellcolor{orange} 0.850} & \multicolumn{1}{c}{0.902} & \multicolumn{1}{c}{0.863}\\
JAXNeRF+& \cellcolor{orange} 0.845 & \multicolumn{1}{c}{\cellcolor{red} 0.966} & \multicolumn{1}{c}{\cellcolor{yellow} 0.813} & \multicolumn{1}{c}{\cellcolor{red} 0.724} & \multicolumn{1}{c}{\cellcolor{orange} 0.900} & \multicolumn{1}{c}{\cellcolor{red} 0.669} & \multicolumn{1}{c}{\cellcolor{red} 0.868} & \multicolumn{1}{c}{\cellcolor{red} 0.921} & \multicolumn{1}{c}{\cellcolor{red} 0.898}\\ \hline 
JAXNeRF+ Deferred& \cellcolor{yellow} 0.837 & \multicolumn{1}{c}{\cellcolor{yellow} 0.957} & \multicolumn{1}{c}{\cellcolor{orange} 0.816} & \multicolumn{1}{c}{\cellcolor{orange} 0.720} & \multicolumn{1}{c}{\cellcolor{red} 0.901} & \multicolumn{1}{c}{\cellcolor{yellow} 0.657} & \multicolumn{1}{c}{\cellcolor{yellow} 0.844} & \multicolumn{1}{c}{\cellcolor{yellow} 0.913} & \multicolumn{1}{c}{\cellcolor{orange} 0.886}\\
SNeRG (PNG)& 0.818 & \multicolumn{1}{c}{0.936} & \multicolumn{1}{c}{0.802} & \multicolumn{1}{c}{0.696} & \multicolumn{1}{c}{0.889} & \multicolumn{1}{c}{0.655} & \multicolumn{1}{c}{0.835} & \multicolumn{1}{c}{0.882} & \multicolumn{1}{c}{0.852}\\
JAXNeRF+ Diffuse& 0.832 & \multicolumn{1}{c}{0.947} & \multicolumn{1}{c}{\cellcolor{red} 0.821} & \multicolumn{1}{c}{0.715} & \multicolumn{1}{c}{0.891} & \multicolumn{1}{c}{0.656} & \multicolumn{1}{c}{0.827} & \multicolumn{1}{c}{\cellcolor{orange} 0.916} & \multicolumn{1}{c}{\cellcolor{yellow} 0.883}\\
\end{tabular}
\caption{SSIM, Real Forward-Facing scenes.}
\label{tbl:per_scene_real_ssim}
\end{table*}

\begin{table*}[]
\centering
\begin{tabular}{l|ccccccccc}
& \multicolumn{9}{c}{LPIPS $\downarrow$} \\
        & Mean & \scenename{Room} & \scenename{Fern} & \scenename{Leaves} & \scenename{Fortress} & \scenename{Orchids} & \scenename{Flower} & \scenename{T-Rex} & \scenename{Horns} \\ \hline 
LLFF& 0.212 & \multicolumn{1}{c}{0.155} & \multicolumn{1}{c}{0.247} & \multicolumn{1}{c}{\cellcolor{red} 0.216} & \multicolumn{1}{c}{0.173} & \multicolumn{1}{c}{0.313} & \multicolumn{1}{c}{0.174} & \multicolumn{1}{c}{0.222} & \multicolumn{1}{c}{0.193}\\
DeRF& 0.274 & \multicolumn{1}{c}{0.160} & \multicolumn{1}{c}{0.300} & \multicolumn{1}{c}{0.310} & \multicolumn{1}{c}{0.320} & \multicolumn{1}{c}{0.340} & \multicolumn{1}{c}{0.240} & \multicolumn{1}{c}{0.220} & \multicolumn{1}{c}{0.300}\\
NeRF& 0.250 & \multicolumn{1}{c}{0.178} & \multicolumn{1}{c}{0.280} & \multicolumn{1}{c}{0.316} & \multicolumn{1}{c}{0.171} & \multicolumn{1}{c}{0.321} & \multicolumn{1}{c}{0.219} & \multicolumn{1}{c}{0.249} & \multicolumn{1}{c}{0.268}\\
IBRNet& 0.175 & --- & --- & --- & --- & --- & --- & --- & ---\\
JAXNeRF& 0.173 & \multicolumn{1}{c}{\cellcolor{orange} 0.086} & \multicolumn{1}{c}{0.207} & \multicolumn{1}{c}{0.247} & \multicolumn{1}{c}{0.108} & \multicolumn{1}{c}{0.266} & \multicolumn{1}{c}{\cellcolor{yellow} 0.156} & \multicolumn{1}{c}{0.143} & \multicolumn{1}{c}{0.173}\\
JAXNeRF+& \cellcolor{red} 0.145 & \multicolumn{1}{c}{\cellcolor{red} 0.066} & \multicolumn{1}{c}{\cellcolor{red} 0.176} & \multicolumn{1}{c}{\cellcolor{orange} 0.233} & \multicolumn{1}{c}{\cellcolor{orange} 0.097} & \multicolumn{1}{c}{\cellcolor{red} 0.238} & \multicolumn{1}{c}{\cellcolor{red} 0.124} & \multicolumn{1}{c}{\cellcolor{red} 0.113} & \multicolumn{1}{c}{\cellcolor{red} 0.119}\\ \hline 
JAXNeRF+ Deferred& \cellcolor{orange} 0.160 & \multicolumn{1}{c}{\cellcolor{yellow} 0.090} & \multicolumn{1}{c}{\cellcolor{yellow} 0.186} & \multicolumn{1}{c}{\cellcolor{yellow} 0.240} & \multicolumn{1}{c}{\cellcolor{red} 0.097} & \multicolumn{1}{c}{0.256} & \multicolumn{1}{c}{\cellcolor{orange} 0.149} & \multicolumn{1}{c}{\cellcolor{yellow} 0.125} & \multicolumn{1}{c}{\cellcolor{yellow} 0.134}\\
SNeRG (PNG)& 0.183 & \multicolumn{1}{c}{0.133} & \multicolumn{1}{c}{0.198} & \multicolumn{1}{c}{0.252} & \multicolumn{1}{c}{0.125} & \multicolumn{1}{c}{\cellcolor{yellow} 0.255} & \multicolumn{1}{c}{0.167} & \multicolumn{1}{c}{0.157} & \multicolumn{1}{c}{0.176}\\
JAXNeRF+ Diffuse& \cellcolor{yellow} 0.164 & \multicolumn{1}{c}{0.115} & \multicolumn{1}{c}{\cellcolor{orange} 0.182} & \multicolumn{1}{c}{0.241} & \multicolumn{1}{c}{\cellcolor{yellow} 0.105} & \multicolumn{1}{c}{\cellcolor{orange} 0.251} & \multicolumn{1}{c}{0.166} & \multicolumn{1}{c}{\cellcolor{orange} 0.116} & \multicolumn{1}{c}{\cellcolor{orange} 0.133}\\
\end{tabular}
\caption{LPIPS, Real Forward-Facing scenes.}
\label{tbl:per_scene_real_lpips}
\end{table*}

\begin{table*}[]
\centering
\begin{tabular}{r|ccc|ccccccccc}
        & MLP & $\mathcal{L}_s$ & Defer & Mean & \scenename{Chair} & \scenename{Drums} & \scenename{Ficus} & \scenename{Hotdog} & \scenename{Lego} & \scenename{Materials} & \scenename{Mic} & \scenename{Ship} \\ \hline 
Ours &\cmark &\cmark &\cmark &  11.9 & 9.8 & 9.6 & 21.1 & 7.2 & 9.3 & 10.1 & 10.1 & 17.9\\
1) &\xmark &\cmark &\xmark &   9.2 & 6.6 & 6.6 & 19.4 & 5.3 & 6.0 & 7.8 & 7.6 & 14.0\\
2) & \cmark  &\xmark &\cmark &  {---} &  15.3 & 15.7 & --- & 24.4 & 25.4 & 19.7 & 12.6 & 27.0\\
3) & \cmark  &\cmark &\xmark &   343.6 & 269.8 & 268.9 & 987.9 & 155.8 & 261.4 & 262.2 & 222.6 & 320.4\\
\end{tabular}
\caption{Performance ablation (milliseconds/frame $\downarrow$), Synthetic $360^\circ$ Scenes.}
\label{tbl:per_scene_time}
\end{table*}

\begin{table*}[]
\centering
\begin{tabular}{r|ccc|ccccccccc}
        & MLP & $\mathcal{L}_s$ & Defer & Mean & \scenename{Chair} & \scenename{Drums} & \scenename{Ficus} & \scenename{Hotdog} & \scenename{Lego} & \scenename{Materials} & \scenename{Mic} & \scenename{Ship} \\ \hline 
Ours &\cmark &\cmark &\cmark & 1.73 & 0.78 & 0.86 & 4.99 & 0.53 & 0.74 & 0.98 & 1.18 & 3.78 \\
1) &\xmark &\cmark &\xmark &  1.73 & 0.78 & 0.86 & 4.99 & 0.53 & 0.74 & 0.98 & 1.18 & 3.78 \\
2) & \cmark  &\xmark &\cmark & 4.26 & 2.44 & 3.25 & 7.57 & 4.17 & 3.68 & 4.91 & 2.33 & 5.77 \\
3) & \cmark  &\cmark &\xmark &  1.73 & 0.78 & 0.86 & 4.99 & 0.53 & 0.74 & 0.98 & 1.18 & 3.78 \\
\end{tabular}
\caption{Performance ablation (GPU Memory in GB $\downarrow$), Synthetic $360^\circ$ Scenes.}
\label{tbl:per_scene_mem}
\end{table*}



\end{document}